\documentclass{article}

\usepackage{microtype}
\usepackage{graphicx}
\usepackage{subfigure}
\usepackage{booktabs} 
\usepackage{multirow}

\usepackage{hyperref}
\usepackage{enumitem}

\usepackage[accepted]{icml2024}

\usepackage{amsmath}
\usepackage{amssymb}
\usepackage{mathtools}
\usepackage{amsthm}

\usepackage[capitalize,noabbrev]{cleveref}

\theoremstyle{plain}
\newtheorem{theorem}{Theorem}[section]

\theoremstyle{definition}
\newtheorem{definition}[theorem]{Definition}

\theoremstyle{remark}

\usepackage[textsize=tiny]{todonotes}

\icmltitlerunning{Unsupervised Episode Generation for Graph Meta-learning}

\begin{document}

\twocolumn[
\icmltitle{Unsupervised Episode Generation for Graph Meta-learning}



\icmlsetsymbol{equal}{*}

\begin{icmlauthorlist}
\icmlauthor{Jihyeong Jung}{isyse}
\icmlauthor{Sangwoo Seo}{isyse}
\icmlauthor{Sungwon Kim}{gsds}
\icmlauthor{Chanyoung Park}{isyse,gsds}

\end{icmlauthorlist}

\icmlaffiliation{isyse}{Department of Industrial \& Systems Engineering, KAIST}
\icmlaffiliation{gsds}{Graduate School of Data Science, KAIST}

\icmlcorrespondingauthor{Chanyoung Park}{cy.park@kaist.ac.kr}

\icmlkeywords{Unsupervised Learning, Graph Neural Networks, Few-Shot Node Classification, Meta-learning}

\vskip 0.3in
]



\printAffiliationsAndNotice{}  

\begin{abstract}
We propose Unsupervised Episode Generation method called \textbf{Neighbors as Queries (\textsc{NaQ})} to solve the Few-Shot Node-Classification (FSNC) task by \textit{unsupervised Graph Meta-learning}.
Doing so enables full utilization of the information of all nodes in a graph, which is not possible in current supervised meta-learning methods for FSNC due to the label-scarcity problem.
In addition, unlike unsupervised Graph Contrastive Learning (GCL) methods that overlook the downstream task to be solved at the training phase resulting in vulnerability to class imbalance of a graph, we adopt the episodic learning framework that allows the model to be aware of the downstream task format, i.e., FSNC.
The proposed \textsc{NaQ} is a simple but effective \textit{unsupervised} episode generation method that randomly samples nodes from a graph to make a support set, followed by similarity-based sampling of nodes to make the corresponding query set.
Since \textsc{NaQ} is \textit{model-agnostic}, any existing supervised graph meta-learning methods can be trained in an unsupervised manner, while not sacrificing much of their performance or sometimes even improving them.
Extensive experimental results demonstrate the effectiveness of our proposed unsupervised episode generation method for graph meta-learning towards the FSNC task.
Our code is available at: \url{https://github.com/JhngJng/NaQ-PyTorch}. 
\end{abstract}

\section{Introduction} \label{intro}
Graph-structured data are useful and widely applicable in the real-world, thanks to their capability of modeling complex relationships between objects such as user-user relationships in social networks and product networks, etc. To handle tasks such as node classification on graph-structured data, Graph Neural Networks (GNNs) are widely used and have shown remarkable performance~\cite{gcn, gat}. 
However, it is well known that GNNs suffer from poor generalization when only a small number of labeled samples are provided~\cite{metagnn, gpn, tent}.

To mitigate such issues inherent in the ordinary deep neural networks, few-shot learning methods have emerged, and the dominant paradigm was applying meta-learning algorithms such as MAML~\cite{maml} and ProtoNet~\cite{protonet}, which are based on the episodic learning framework~\cite{vinyals2016matching}. Inspired by these methods, recent studies proposed graph meta-learning methods~\cite{metagnn, gpn, gmeta, tent} to solve the Few-Shot Node Classification (FSNC) task on graphs by also leveraging the episodic learning framework, which is the main focus of this study.

\begin{figure*} [t]
    \centering
    \subfigure[Impact of the label-scarcity]{
        \includegraphics[width=0.25\textwidth]{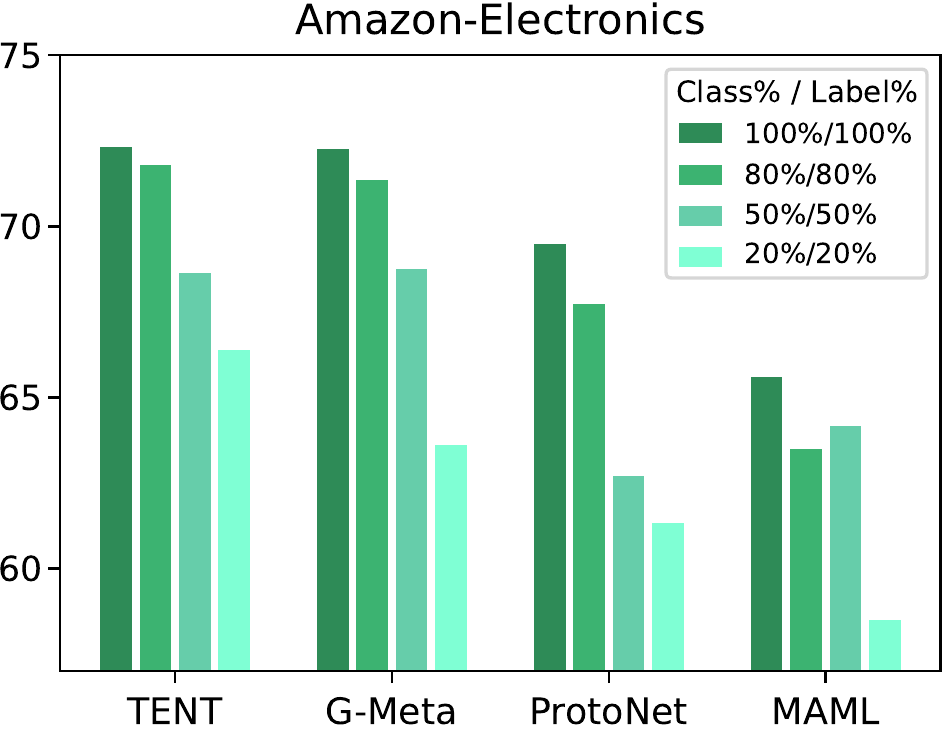}
        \label{intro:sup_label}
    }
    \hspace{2mm}
    \subfigure[Impact of the label noise]{
        \includegraphics[width=0.27\textwidth]{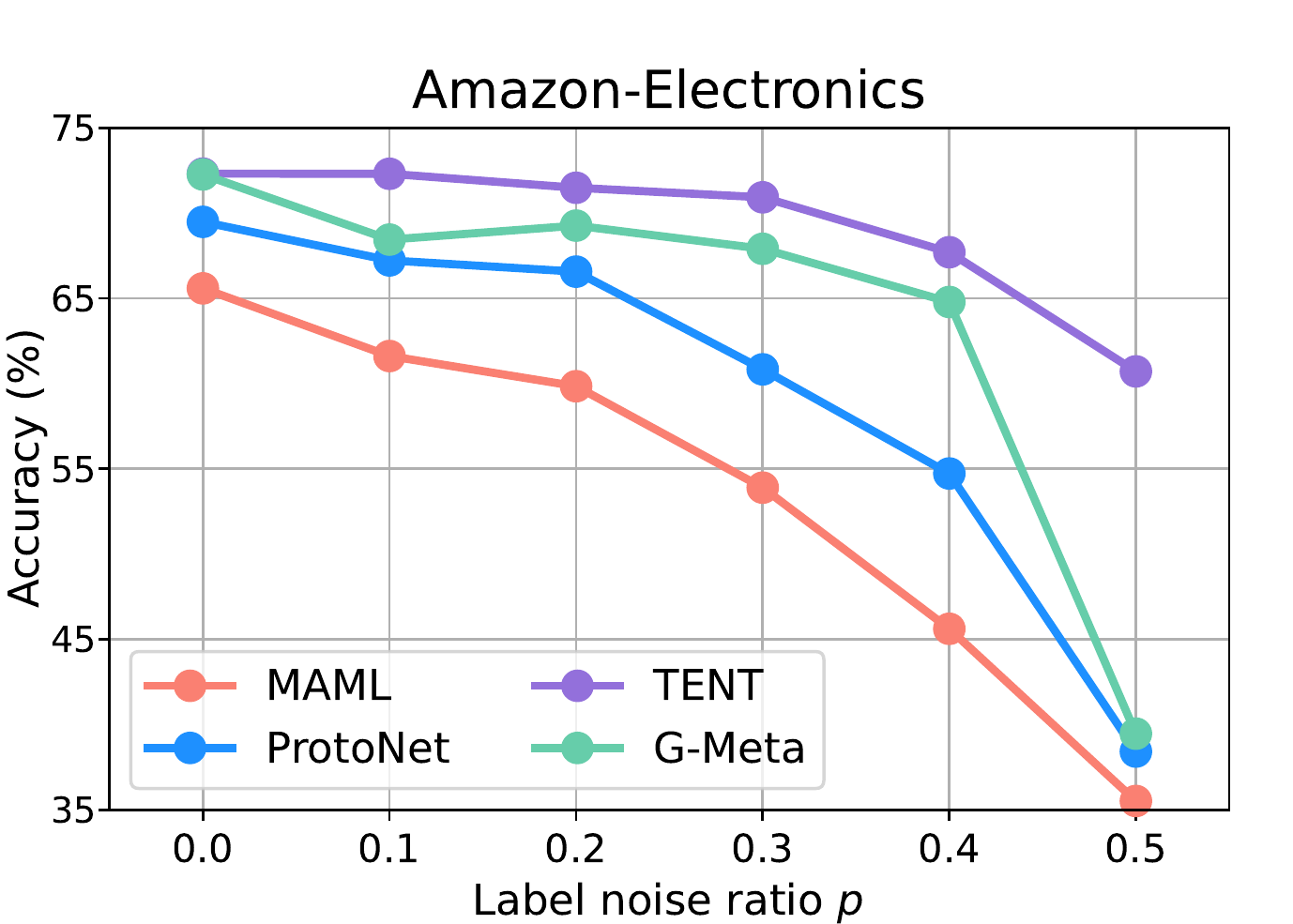}
        \label{intro:label_noise}
    }
    \hspace{2mm}
    \subfigure[Impact of the class imbalance]{
        \includegraphics[width=0.26\textwidth]{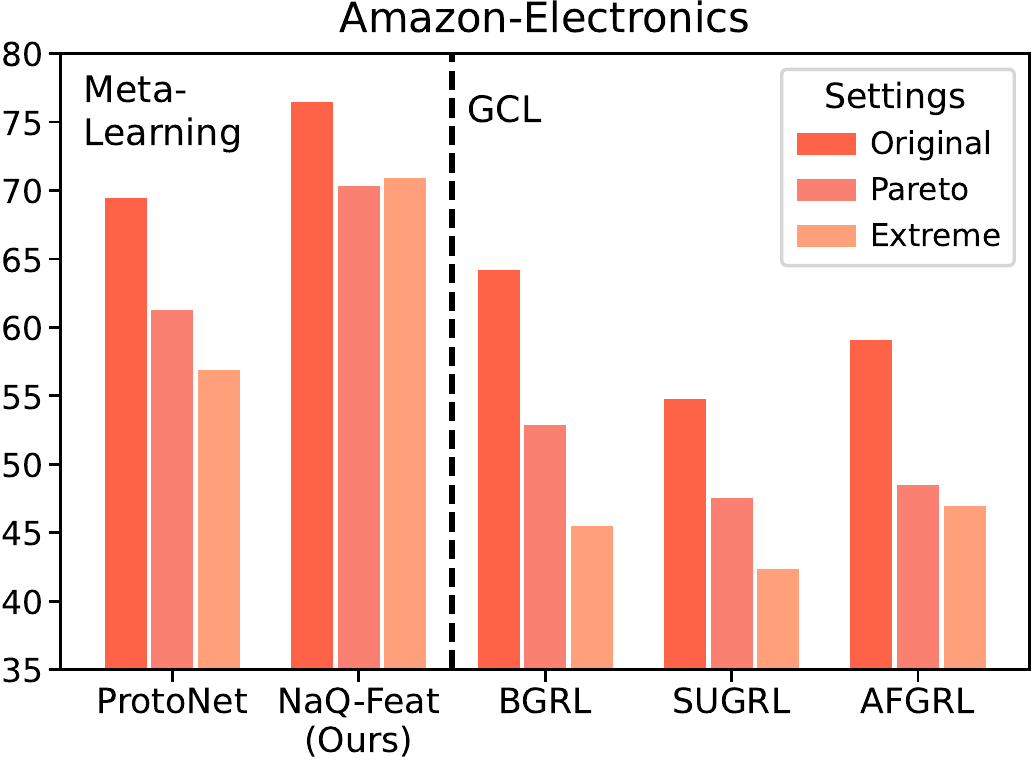}
        \label{intro:class_imbal}
    }
    \vspace{-2ex}
    \caption{(a): Impact of the label-scarcity on supervised graph meta-learning methods (`Class\%': a rate of available base classes during training, `Label\%': a rate of available labeled samples for each class). (b): Impact of the (randomly injected) label noise $p$ on supervised graph meta-learning methods. (c): Impact of the class imbalance (`Pareto' setting: we kept nodes for top-20\% head classes, while keeping only 10 nodes for remaining classes; `Extreme' setting: the only difference from the `Pareto' setting is that we kept nodes only for top-5 head classes instead of top-20\% classes). (5-way 1-shot)}
    \label{intro:fig}
    \vspace{-2ex}
\end{figure*}

Despite their effectiveness, existing supervised graph meta-learning methods require \textit{abundant} labeled samples from \textit{diverse} base classes for the training.
As shown in Figure~\ref{intro:sup_label}, such label-scarcity causes a severe performance drop of representative methods (i.e., TENT~\cite{tent}, G-Meta~\cite{gmeta}, ProtoNet~\cite{protonet}, and MAML~\cite{maml}) in FSNC.
However, gathering enough labeled data and diverse classes may not be possible, and is costly in reality.
More importantly, as these methods depend on a few labeled nodes from base classes, while not fully utilizing all nodes in the graph, they are also vulnerable to noisy labels in base classes (Figure~\ref{intro:label_noise}).
In this respect, \textit{unsupervised methods are indispensable to fundamentally address the label-dependence problem} of existing supervised graph meta-learning methods.

Most recently, TLP~\cite{tlp} empirically demonstrated that a simple linear probing with node embeddings pre-trained by Graph Contrastive Learning (GCL) methods outperforms existing supervised graph meta-learning methods in FSNC. This is because GCL methods tend to generate generic node embeddings, since all nodes in a graph are involved in the training.

However, despite the effectiveness of generic node embeddings, we argue that \textit{they are vulnerable to class imbalance in the graph}, which might lead to a significant performance drop due to the lack of model generalizability resulting from the discrepancy in the objective between pre-training and fine-tuning (in downstream task) phase~\cite{lu2021learning}.
If the given graph mainly consists of nodes from the majority classes, GCL methods have difficulty in learning embeddings of nodes from the minority classes, which results in poor FSNC performance on such minority classes.
On the other hand, as each episode in the episodic learning framework {provides the GNN encoder with the information about the downstream task format (i.e., FSNC), meta-learning methods are rather more robust to the class imbalance that may exist in the given graph\footnote{Please refer to Section~\ref{imbal:discuss} for more detail regarding how the episodic learning allows the model to be robust to class imbalance.}.
To corroborate our argument, we modified the original graph to simulate two class imbalance settings (i.e., `Pareto' and `Extreme'), and evaluated GCL methods (i.e., BGRL~\cite{bgrl}, SUGRL~\cite{sugrl}, AFGRL~\cite{afgrl}) and meta-learning methods (i.e., ProtoNet and \textsc{NaQ-Feat} (ours)) on the FSNC task (See Figure~\ref{intro:class_imbal}).
As expected, the performance deterioration of GCL methods was more severe than meta-learning methods under class imbalance settings.

Therefore, we argue that the FSNC performance can be further enhanced by \textbf{unsupervised Graph Meta-learning}, \textit{which can achieve the best of both worlds}: 1) GCL that fully utilizes all nodes in a graph in an unsupervised manner, and 2) Meta-learning whose episodic learning framework is aware of the downstream task format (i.e., \textit{FSNC}).

In this work, we propose a simple yet effective \textit{unsupervised} episode generation method called \textbf{Neighbors as Queries (\textsc{NaQ})}, which enables unsupervised graph meta-learning, to benefit from the generalization ability of meta-learning methods for the FSNC task, while fully utilizing all nodes in a graph. 
The main idea is to construct a support set by randomly choosing nodes from the entire graph, and generate a corresponding query set via sampling similar nodes based on pre-calculated node-node similarity. 
It is important to note that our unsupervised episode generation method is \textit{model-agnostic}, i.e., \textsc{NaQ} can be used to train any existing supervised graph meta-learning methods in an unsupervised manner directly or only with minor modifications.

To sum up, our contributions are summarized as follows: \vspace{-2ex}
\begin{enumerate}[leftmargin=0.4cm]
    \item We present an \textit{unsupervised episode generation} method, called \textsc{NaQ}, designed to solve the FSNC task via unsupervised graph meta-learning.
    To our best knowledge, this is the first study that focuses on the unsupervised episode generation of graph meta-learning framework. \vspace{-1ex}
    \item \textsc{NaQ} is \textit{model-agnostic}; that is, it can be used to train any existing supervised graph meta-learning methods in an unsupervised manner, while not sacrificing much of their performance or sometimes even improving them, without using any labeled nodes. \vspace{-1ex}
    \item Extensive experimental results demonstrate the effectiveness of \textsc{NaQ} in the FSNC task and highlight the potential of the \textit{unsupervised} graph meta-learning framework.
\end{enumerate}

\section{Preliminaries}
\subsection{Problem Statement}
Let $\mathcal{G}=(\mathcal{V}, \mathcal{E}, X)$ be a graph, where $\mathcal{V},\  \mathcal{E}\subset\mathcal{V}\times\mathcal{V},\ X\in\mathbb{R}^{|\mathcal{V}|\times d}$ are a set of nodes, a set of edges, and a $d$-dimensional node feature matrix, respectively. We also use $X$ to denote a set of node features, i.e., $X=\{x_{v}:v\in \mathcal{V}\}$.
Let $C$ be a set of total node classes. Here, we denote the \textit{base classes}, a set of node classes that can be utilized during training, as $C_b$, and denote the \textit{target classes}, a set of node classes that we aim to predict in downstream tasks given a few labeled samples, as $C_t$. Note that $C_b\cup C_t=C$ and $C_b\cap C_t=\emptyset$, and the target classes $C_t$ are unknown during training.
In common few-shot learning settings, the number of labeled nodes from classes of $C_b$ is sufficient, while we only have a few labeled nodes from classes of $C_t$ in downstream tasks. 
Now we formulate the ordinary supervised few-shot node classification (FSNC) problem as follows:
\begin{definition} [Supervised FSNC]
    Given a graph $\mathcal{G}=(\mathcal{V}, \mathcal{E}, X)$, labeled data $(X_{C_b}, Y_{C_b})$ and a model $f_{\theta}$ trained on $(X_{C_b}, Y_{C_b})$, the goal of supervised FSNC is making predictions for $x_{q}\in X_{C_t}$ ({i.e. \textit{query set}}) based on a few labeled samples $(x_{s}, y_{s})\in (X_{C_t}, Y_{C_t})$ ({i.e., \textit{support set}}) during the testing phase.
\end{definition}
Based on this problem formulation, we can formulate the unsupervised FSNC problem as below.
The only difference is that labeled nodes are not available during training.
\begin{definition} [Unsupervised FSNC] 
    Given a graph $\mathcal{G}=(\mathcal{V}, \mathcal{E}, X)$, \textbf{unlabeled data} $X=X_{C_b}\cup X_{C_t}$, and a model $f_\theta$ trained on $X$, the goal of unsupervised FSNC is making predictions for $x_{q}\in X_{C_t}$ ({i.e., \textit{query set}}) based on a few labeled samples $(x_{s}, y_{s})\in (X_{C_t}, Y_{C_t})$ ({i.e., \textit{support set}}) during the testing phase.
\end{definition}
Overall, the goal of FSNC is to adapt well to unseen target classes $C_t$ only using a few labeled samples from $C_t$ after training a model $f_\theta$ on training data.
In this work, we study how to facilitate \textit{unsupervised Graph Meta-learning} to solve the FSNC task.
More formally, we consider solving a $N$-way $K$-shot FSNC task~\cite{vinyals2016matching}, where $N$ is the number of distinct target classes and $K$ is the number of labeled samples in a support set. Moreover, there are $Q$ query samples to be classified in each downstream task.

\subsection{Episodic Learning Framework}
We follow the episodic training framework~\cite{vinyals2016matching} that is formally defined as follows: 
\begin{definition} [Episodic Learning]
    Episodic learning is a learning framework that utilizes a bundle of tasks $\{\mathcal{T}_{t}\}_{t=1}^{T}$, where $\mathcal{T}_{t}=(S_{\mathcal{T}_{t}},\  Q_{\mathcal{T}_{t}})$, $S_{\mathcal{T}_{t}}=\{ (x^{spt}_{t,i}, y^{spt}_{t,i}) \}_{i=1}^{N\times K}$ and $Q_{\mathcal{T}_{t}}=\{ (x^{qry}_{t,i}, y^{qry}_{t,i}) \}_{i=1}^{N\times Q}$, instead of commonly used mini-batches in the stochastic optimization.
\end{definition}
\noindent By mimicking the `format' of the downstream task ({i.e., \textit{FSNC}), the episodic learning allows the model to be aware of the task to be solved in the testing phase.
Note that existing supervised meta-learning methods require a large number of labeled samples in the training set $(X_{C_b}, Y_{C_b})$ and a sufficient number of base classes $|C_b|$ ({i.e., \textit{diverse base classes}}) to generate informative training episodes.
However, gathering enough labeled data and diverse classes may not be possible and is usually costly in the real world.
As a result, supervised methods fall short of utilizing all nodes in the graph as they rely on a few labeled nodes, and thus lack generalizability.

Therefore, we propose \textit{unsupervised episode generation} methods not only to tackle the label-scarcity problem causing a limited utilization of nodes in the graph, but also to benefit from the episodic learning framework for downstream task-aware learning of node embeddings, thereby being robust to a class imbalance in a graph.

\section{Proposed Method} 
\subsection{Motivation: A Closer Look at Training Episodes} \label{motivation}
In the episodic learning framework, there are two essential components: 1) support set that provides basic information about the task to be solved, and 2) query set that enables the model to understand about how to solve the given task.
\textit{For this reason, the query set should share similar semantics with the support set.}
Motivated by this characteristic of episodic learning, we consider the \textbf{similarity} condition as the key to our proposed query generation process.
Note that in the ordinary supervised setting, the similarity condition is easily achievable, since labels of the support set and query set are known, and thus can be sampled from the same class. However, as our goal is to generate training episodes in an unsupervised manner, how to sample a query set that shares similar semantics with each support set is non-trivial.

\begin{figure} [t!]
  \centering
  \includegraphics[width=0.99\columnwidth]{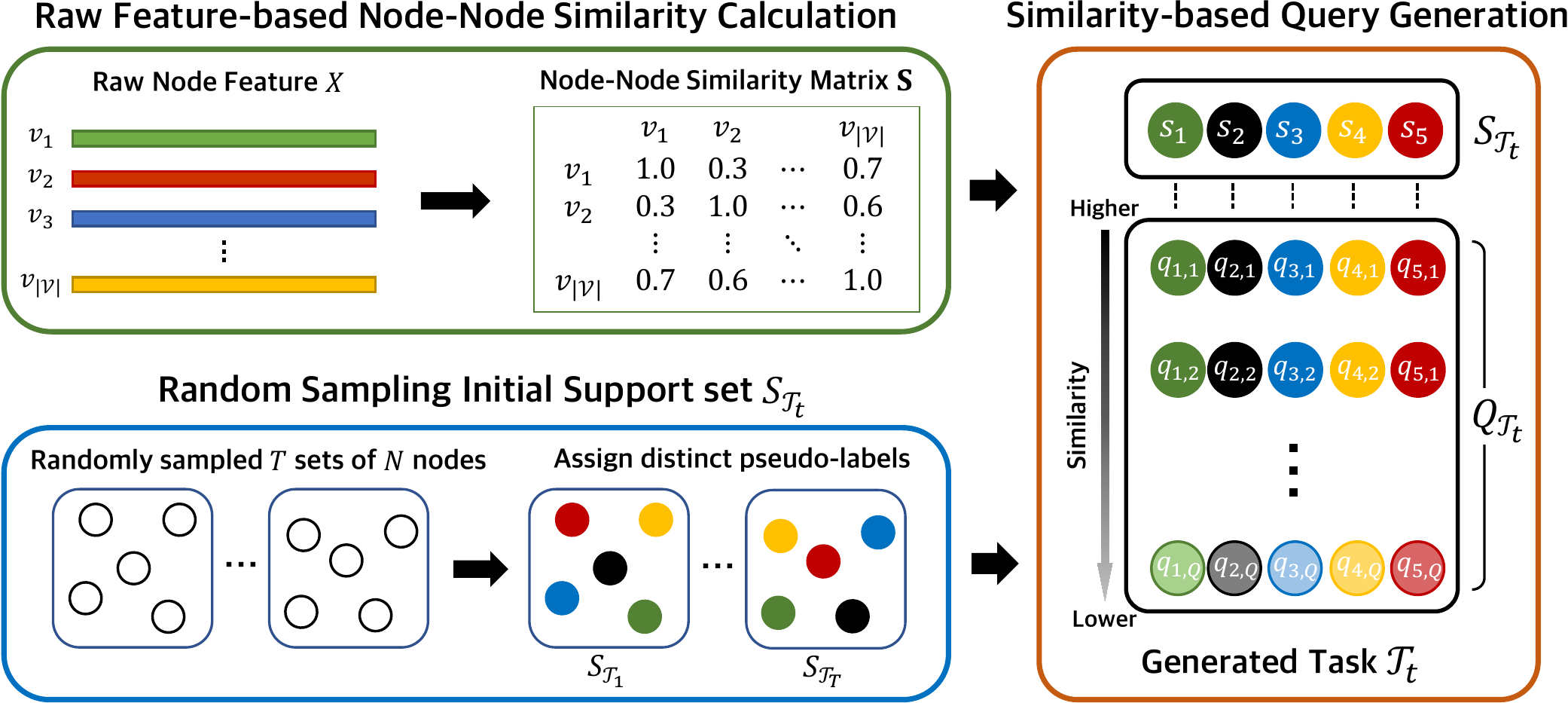}
  \vspace{-3mm}
  \caption{Overview of the \textsc{NaQ-Feat}.}
  \label{naq_feat:overview}
  \vspace{-1ex}
\end{figure}

\subsection{\textsc{NaQ}: Neighbors as Queries} \label{naq:explain}
In this work, we propose a simple yet effective query generation method, called \textbf{Neighbors as Queries (\textsc{NaQ})}, which leverages raw feature-level similar nodes as queries. The overview of \textsc{NaQ} can be found in Figure~\ref{naq_feat:overview}.

\noindent \textbf{Support set generation. }
{To generate training episodes $\{\mathcal{T}_{t}\}_{t=1}^{T}$, we start by randomly sampling $T$ sets of $N$ nodes from the entire graph for the support set generation. 
Next, we assign pseudo-labels $y_{t,i}$ to each node $x_{t,i} \in \mathcal{T}_t$, {i.e.}, $S_{\mathcal{T}_t}=\{ (x_{t,i}, y_{t,i}) \ | \ x_{t,i}\in \mathcal{V} \}_{i=1}^{N\times K}$.}
Note that we only generate 1-shot support set (i.e., $K=1$) regardless of the downstream task setting, to assure that randomly sampled $N$ support set nodes (corresponding to `$N$-way') are as much distinguishable from one another as possible.

\noindent \textbf{Query set generation. }
Then, we generate a corresponding query set $Q_{\mathcal{T}_t}$ with Top-Q similar nodes of each node $x_{t,i}$ in $S_{\mathcal{T}_t}$ based on a pre-calculated node-node similarity matrix $\mathbf{S}$, and give them the same pseudo-label $y_{t,i}$.
Formally, we can express this query generation process as follows:
\begin{equation}
    Q_{\mathcal{T}_t} = \bigcup\nolimits_{(x_{t,i}, y_{t,i}) \in S_{\mathcal{T}_t}} \mathrm{Top}(\mathbf{S}_{x_{t,i}}, \text{Q})
\end{equation}
where $\mathbf{S}_{x_{t,i}}$ denotes a row of the similarity matrix $\mathbf{S}$ corresponding to the node $x_{t,i}$, and $\mathrm{Top}(\mathbf{S}_{x_{t,i}}, \text{Q})$ indicates a set of Q nodes corresponding to Q largest entries in $\mathbf{S}_{x_{t,i}}$ excluding $x_{t,i}$ itself.

\noindent\textbf{Similarity Metric. } For sampling `similar' nodes to be used as queries, we used cosine similarity for node features such as bag-of-words, and Euclidean distance for the features such as word embeddings. Refer to Section~\ref{sim:measure} in the Appendix for further discussions on the similarity metric.

\subsubsection{An Extension to \textsc{NaQ}: \textsc{NaQ-Diff}} \label{naq_diff:rise}
Since \textsc{NaQ} described above solely relies on the raw node feature $X$, the structural information that is inherent in graphs is overlooked, which plays an important role depending on the target domain.
For example, in citation networks, since the citation relationship between papers implies that these papers usually share similar semantics ({i.e., related paper topics}), they have similar features even if their class labels are different.
Hence, considering structurally similar nodes as queries can be more beneficial than solely relying on the feature-level similar nodes in such cases.

Hence, we present a variant of \textsc{NaQ}, called \textsc{NaQ-Diff}, which utilizes \textit{structurally similar nodes} found by generalized graph diffusion~\cite{gdc} as queries. Specifically,
\textsc{NaQ-Diff} leverages diffusion matrix $\mathbf{S}=\Sigma^\infty_{k=0}\theta_k\mathbf{T}^k$ as node-node similarity matrix, with weighting coefficients $\theta_k$, and the generalized transition matrix $\mathbf{T}$. As edge weights of the diffusion matrix $\mathbf{S}$ can be interpreted as structural closeness, we can sample similar nodes of each support set node from $\mathbf{S}$. 
It is important to note that computing the diffusion matrix does not require additional computation during training and can be readily calculated before the model training.
The overview of \textsc{NaQ-Diff} can be found in Figure~\ref{naq_diff:overview} in the Appendix, and detailed settings for \textsc{NaQ-Diff} can be found in Section~\ref{naq_diff}. 
Hereafter, we call the former version of \textsc{NaQ} that is based on the raw features as \textsc{NaQ-Feat}, and the latter version that is based on the graph structural information as \textsc{NaQ-Diff}.

\subsection{Model Training with Episodes from \textsc{NaQ}} \label{sec:training}
In this section, we explain how to train existing meta-learning models with episodes generated by \textsc{NaQ}.
Let $\mathcal{T}_t=(S_{\mathcal{T}_t}, Q_{\mathcal{T}_t})$ be a generated episode and $\mathrm{Meta}(\mathcal{T}_t;\theta)$ be any of existing graph meta-learning methods (e.g., MAML, ProtoNet, G-Meta, etc.) with parameter $\theta$. For simplicity of explanation, we used the same notation here even for methods that use meta-batches like MAML. Regardless of whether $\mathcal{T}_t$ is generated from \textsc{NaQ} or an ordinary supervised episode generation, it follows the common format of $\mathcal{T}_t=(S_{\mathcal{T}_t}, Q_{\mathcal{T}_t}), \text{where } S_{\mathcal{T}_t}=\{(x^{spt}_{t,i}, y^{spt}_{t,i})\}_{i=1}^{N\times K} \text{ and } Q_{\mathcal{T}_t}=\{(x^{qry}_{t,i}, y^{qry}_{t,i})\}_{i=1}^{N\times Q}$. That is, the only difference is whether $y^{spt}_{t,i}$ and $y^{qry}_{t,i}$ are annotated based on the ground-truth label (supervised) or a psuedo-label (\textsc{NaQ}).
Hence, \textit{any of $\mathrm{Meta}(\mathcal{T}_t;\theta)$ can be trained in the same way in an `unsupervised manner with \textsc{NaQ}' as ordinary supervised meta-learning methods.} The details are presented in Algorithm~\ref{alg:naq:train}.

\begin{algorithm} [h!]
\caption{Training Meta-learner $\mathrm{Meta}( \ \cdot \ ;\theta)$ with \textsc{NaQ}}
\label{alg:naq:train}
\begin{algorithmic}
    \INPUT Bundle of training episodes $\{\mathcal{T}_t\}_{t=1}^{T}$, Graph Meta-learner $\mathrm{Meta}( \ \cdot \ ;\theta)$, learning rate $\eta$.
    \STATE Randomly initialize the model parameter $\theta$
    \FOR{$t=1,\cdots,T$}
    \STATE {\bfseries Step 1:} Calculate loss $\mathcal{L}$ by $\mathrm{Meta}(\mathcal{T}_t;\theta)$
    \STATE {\bfseries Step 2:} Update $\theta \leftarrow \theta - \eta \nabla_{\theta} \mathcal{L}$
    \ENDFOR
    \OUTPUT $\mathrm{Meta}(\mathcal{T}_t;\theta)$
\end{algorithmic}
\end{algorithm}

\noindent \textbf{Remark.} Supervised TENT~\cite{tent} additionally computes cross-entropy loss $\mathcal{L}_{CE}$ over the entire labeled data $(X_{C_b}, Y_{C_b})$ in Step 1 of Algorithm~\ref{alg:naq:train}. Therefore, when we train TENT with our \textsc{NaQ}, $\mathcal{L}_{CE}$ is calculated over a single training episode.

It is important to note that since \textsc{NaQ} generates training episodes based on all nodes in a graph, it enables existing graph meta-learning methods to fully utilize all nodes in a graph, while the supervised episode generation fails to do so as it depends on a few labeled nodes from base classes. The detailed model training example in case of ProtoNet~\cite{protonet} can be found in Section~\ref{appx:training:proto} in the Appendix.

\begin{table*}[t!]
\centering
\caption{Overall averaged FSNC accuracy (\%) with 95\% confidence intervals on product networks (Full ver. available at: Table~\ref{main_table:full:acae})}
\vspace{1mm}
\label{main_table:acae}
\renewcommand{\arraystretch}{1.1}
\resizebox{0.95\textwidth}{!}{%
\begin{tabular}{c|ccccc|ccccccc}
\hline
Dataset & \multicolumn{5}{c|}{\textbf{Amazon-Clothing}} & \multicolumn{7}{c}{\textbf{Amazon- Electronics}} \\ \hline
Setting & \multicolumn{2}{c}{5 way} & \multicolumn{2}{c|}{10 way} & \multirow{2}{*}{\begin{tabular}[c]{@{}c@{}}Avg.\\ Rank\end{tabular}} & \multicolumn{2}{c}{5 way} & \multicolumn{2}{c}{10 way} & \multicolumn{2}{c|}{20 way} & \multirow{2}{*}{\begin{tabular}[c]{@{}c@{}}Avg.\\ Rank\end{tabular}} \\ \cline{1-5} \cline{7-12}
Baselines & 1 shot & 5 shot & 1 shot & \multicolumn{1}{c|}{5 shot} &  & 1 shot & 5 shot & 1 shot & 5 shot & 1 shot & \multicolumn{1}{c|}{5 shot} &  \\ \hline
MAML (Sup.) & 76.13±\tiny{1.17} & 84.28±\tiny{0.87} & 63.77±\tiny{0.83} & \multicolumn{1}{c|}{76.95±\tiny{0.65}} & 10.25 & 65.58±\tiny{1.26} & 78.55±\tiny{0.96} & 57.31±\tiny{0.87} & 67.56±\tiny{0.73} & 46.37±\tiny{0.61} & \multicolumn{1}{c|}{60.04±\tiny{0.52}} & 9.33 \\
ProtoNet (Sup.) & 75.52±\tiny{1.12} & 89.76±\tiny{0.70} & 65.50±\tiny{0.82} & \multicolumn{1}{c|}{82.23±\tiny{0.62}} & 7.25 & 69.48±\tiny{1.22} & 84.81±\tiny{0.82} & 57.67±\tiny{0.85} & 75.79±\tiny{0.67} & 48.41±\tiny{0.57} & \multicolumn{1}{c|}{67.31±\tiny{0.47}} & 5.83 \\
TENT (Sup.) & 79.46±\tiny{1.10} & 89.61±\tiny{0.70} & 69.72±\tiny{0.80} & \multicolumn{1}{c|}{84.74±\tiny{0.59}} & 5.25 & 72.31±\tiny{1.14} & 85.25±\tiny{0.81} & 62.13±\tiny{0.83} & 77.32±\tiny{0.67} & 52.45±\tiny{0.60} & \multicolumn{1}{c|}{69.39±\tiny{0.50}} & 4.00 \\
G-Meta (Sup.) & 78.67±\tiny{1.05} & 88.79±\tiny{0.76} & 65.30±\tiny{0.79} & \multicolumn{1}{c|}{80.97±\tiny{0.59}} & 7.75 & 72.26±\tiny{1.16} & 84.44±\tiny{0.83} & 61.32±\tiny{0.86} & 74.92±\tiny{0.71} & 50.39±\tiny{0.59} & \multicolumn{1}{c|}{65.73±\tiny{0.48}} & 5.67 \\
GLITTER (Sup.) & 75.73±\tiny{1.10} & 89.18±\tiny{0.74} & 64.30±\tiny{0.79} & \multicolumn{1}{c|}{77.73±\tiny{0.68}} & 9.00 & 66.91±\tiny{1.22} & 82.59±\tiny{0.83} & 57.12±\tiny{0.88} & 76.26±\tiny{0.67} & 49.23±\tiny{0.57} & \multicolumn{1}{c|}{61.77±\tiny{0.52}} & 7.00 \\
COSMIC (Sup.) & 82.24±\tiny{0.99} & 91.22±\tiny{0.73} & \underline{74.44±\tiny{0.75}} & \multicolumn{1}{c|}{81.58±\tiny{0.63}} & 3.75 & 72.61±\tiny{1.05} & 86.92±\tiny{0.76} & 65.24±\tiny{0.82} & 78.00±\tiny{0.64} & 58.71±\tiny{0.57} & \multicolumn{1}{c|}{70.29±\tiny{0.44}} & 3.00 \\ \hline
TLP-BGRL & 81.42±\tiny{1.05} & 90.53±\tiny{0.71} & 72.05±\tiny{0.86} & \multicolumn{1}{c|}{83.64±\tiny{0.63}} & 4.25 & 64.20±\tiny{1.10} & 81.72±\tiny{0.85} & 53.16±\tiny{0.82} & 73.70±\tiny{0.66} & 44.57±\tiny{0.54} & \multicolumn{1}{c|}{65.13±\tiny{0.47}} & 8.67 \\
TLP-SUGRL & 63.32±\tiny{1.19} & 86.35±\tiny{0.78} & 54.81±\tiny{0.77} & \multicolumn{1}{c|}{73.10±\tiny{0.63}} & 11.50 & 54.76±\tiny{1.06} & 78.12±\tiny{0.92} & 46.51±\tiny{0.80} & 68.41±\tiny{0.71} & 36.08±\tiny{0.52} & \multicolumn{1}{c|}{57.78±\tiny{0.49}} & 11.67 \\
TLP-AFGRL & 78.12±\tiny{1.13} & 89.82±\tiny{0.73} & 71.12±\tiny{0.81} & \multicolumn{1}{c|}{83.88±\tiny{0.63}} & 5.25 & 59.07±\tiny{1.07} & 81.15±\tiny{0.85} & 50.71±\tiny{0.85} & 73.87±\tiny{0.66} & 43.10±\tiny{0.56} & \multicolumn{1}{c|}{65.44±\tiny{0.48}} & 9.00 \\ \hline
VNT & 65.09±\tiny{1.23} & 85.86±\tiny{0.76} & 62.43±\tiny{0.81} & \multicolumn{1}{c|}{80.87±\tiny{0.63}} & 10.50 & 56.69±\tiny{1.22} & 78.02±\tiny{0.97} & 49.98±\tiny{0.83} & 70.51±\tiny{0.73} & 42.10±\tiny{0.53} & \multicolumn{1}{c|}{60.99±\tiny{0.50}} & 10.83 \\ \hline
\textbf{\textsc{NaQ-Feat}-Best (Ours)} & \textbf{86.58±\tiny{0.96}} & \textbf{92.27±\tiny{0.67}} & \textbf{79.55±\tiny{0.78}} & \multicolumn{1}{c|}{\textbf{86.10±\tiny{0.60}}} & \textbf{1.00} & \textbf{76.46±\tiny{1.11}} & \textbf{88.72±\tiny{0.73}} & \textbf{69.59±\tiny{0.86}} & \textbf{81.44±\tiny{0.61}} & \textbf{61.05±\tiny{0.59}} & \multicolumn{1}{c|}{\textbf{74.60±\tiny{0.47}}} & \textbf{1.00} \\
\textbf{\textsc{NaQ-Diff}-Best (Ours)} & \underline{84.40±\tiny{1.01}} & \underline{91.72±\tiny{0.69}} & 73.39±\tiny{0.79} & \multicolumn{1}{c|}{\underline{84.82±\tiny{0.58}}} & \underline{2.25} & \underline{74.16±\tiny{1.08}} & \underline{87.09±\tiny{0.75}} & \underline{65.95±\tiny{0.81}} & \underline{79.13±\tiny{0.60}} & \underline{60.40±\tiny{0.59}} & \multicolumn{1}{c|}{\underline{73.75±\tiny{0.42}}} & \underline{2.00} \\ \hline
\end{tabular}%
}
\end{table*}

\subsection{Theoretical Insights} \label{theo:insight}
In this section, we provide some insights on conditions that 
enable \textsc{NaQ} to work within the episodic learning framework to justify our motivation of utilizing similar nodes as queries described in Section 3.1.
Specifically, we investigate the learning behavior of MAML~\cite{maml}, which is one of the most widely adopted meta-learning methods in the perspective of `generalization error' for a single episode during the training phase. Since each of the existing graph meta-learning methods has its own sophisticated architecture, we only consider MAML here.
The formal definition of the expected generalization error is as follows~\cite{islr2013, esl}.
\begin{definition} \label{def_generr}
    Let ${S}, {Q}, f_{{S}}, f$ be a given training set, test set, an encoder trained on ${S}$, and the unknown perfect estimation, respectively. With an error measure $\mathcal{L}$, for a given point $(x',y')\in{Q}$, an \textit{expected generalization error} is defined as $\mathbb{E}[\mathcal{L}\big( y', f_{{S}}(x') \big)]$.
\end{definition}
By assuming that $y=f(x)+\epsilon$ holds for an arbitrary input-output pair $(x,y)$ ($\mathbb{E}[\epsilon]=0, \mathrm{Var}(\epsilon)=\sigma^2<\infty$) and an error measure $\mathcal{L}$ is the mean squared error, we can decompose expected generalization error in Def.~\ref{def_generr} as follows~\cite{islr2013, esl}: 

\begin{equation} \label{generr}
\begin{split}
    \mathbb{E}[\mathcal{L}\big( y', f_{{S}}(x')& \big)]  = \big( \mathbb{E} [f_{{S}}(x')] -f(x') \big)^{2} \\ 
    & + \big( \mathbb{E}\big[ f_{{S}}(x')^{2} \big] - \mathbb{E}[f_{{S}}(x')]^{2} \big) + \sigma^{2}.
\end{split}
\end{equation}
Let us consider the training process of MAML with an encoder $f_\theta$ and a training episode $\mathcal{T}=(S_\mathcal{T}, Q_\mathcal{T})$, where $S_\mathcal{T}=\{(x^{spt}_i, y^{spt}_i)\}^{N\times K}_{i=1}~\text{and}~Q_\mathcal{T}$ are the $N$-way $K$-shot support set and the query set, respectively.
During the inner-loop optimization, MAML produces $f_{\theta'}$, where $\theta'=\mathrm{argmin}_{\theta}\sum_{(x^{spt},y^{spt})\in S_\mathcal{T}} \mathcal{L} \big( y^{spt}, f_{\theta}(x^{spt}) \big)$.

If we regard the inner-loop optimization of MAML as a training process with training set $S=S_\mathcal{T}$, the outer-loop optimization (i.e., meta-optimization) as a testing process with test set $Q=Q_\mathcal{T}$, and the trained encoder $f_{S}=f_{S_\mathcal{T}}=f_{\theta'}$, \textit{we can interpret that the meta-optimization actually reduces the generalization error in Eq.~\ref{generr}} over the query set $Q_\mathcal{T}$ with encoder $f_{\theta'}$~\cite{umtra}. With this interpretation, we can re-write Eq.~\ref{generr} as follows:
\begin{equation} \label{generr_meta}
\small
\begin{split}
    \mathbb{E}[\mathcal{L}\big( y^{qry}, f_{\theta'}&(x^{qry}) \big)] = \big( \mathbb{E} [f_{\theta'}(x^{qry})] -f_\mathcal{T}(x^{qry}) \big)^{2} \\
    & + \big( \mathbb{E}\big[ f_{\theta'}(x^{qry})^{2} \big] - \mathbb{E}[f_{\theta'}(x^{qry})]^{2} \big) + \sigma^{2}, 
\end{split}
\end{equation}
where $f_\mathcal{T}$ is the unknown perfect estimation for $\mathcal{T}$. Without loss of generality, we considered a single query $(x^{qry},y^{qry})\in Q_\mathcal{T}$ to derive Eq.~\ref{generr_meta}. \textit{As Eq.~\ref{generr_meta} is used as a loss function, an accurate calculation of Eq.~\ref{generr_meta} is essential for a better model training} on $\mathcal{T}$~\cite{umtra}.

\noindent\textbf{Remark.} Let $s:=(x^{spt},y^{spt})\in S_\mathcal{T}$ be a specific corresponding support set sample of the query $q:=(x^{qry},y^{qry})$ above. Let $\tilde{y}^{spt}$, $\tilde{y}^{qry}$ be the \textit{true labels} of $s, q$, respectively. Note that the \textit{same new labels} (i.e., $y^{spt}$, $y^{qry}$ s.t. $y^{spt}$=$y^{qry}$) are assigned to each of $x^{spt}, x^{qry}$ during the training episode generation (regardless of whether it is supervised or not), to perform classification of $N$ classes instead of classifying $|C|$ classes (i.e., total number of classes in the entire dataset) in the training phase of the episodic learning framework. To get an accurate computation of Eq.~\ref{generr_meta}, \textit{it is essential to assure that $\tilde{y}^{spt}=\tilde{y}^{qry}$ holds}.
Otherwise, we have $y^{qry} = f_{\mathcal{T}}(x^{qry})+\epsilon + \delta$, where $\delta$ is an error resulting from $\tilde{y}^{spt} \neq \tilde{y}^{qry}$, which may lead to a suboptimal solution when training with loss defined by Eq.~\ref{generr_meta}.

Unlike the ordinary supervised episode generation in which case $\delta=0$ holds as condition that $\tilde{y}^{spt}=\tilde{y}^{qry}$ is naturally satisfied, our \textsc{NaQ} cannot guarantee $\delta=0$ since no label information is given (i.e., $\tilde{y}^{spt}, \tilde{y}^{qry}$ are both unknown) during its episode generation phase.
Hence, we argue that it is crucial to discover \textbf{class-level similar} query $q_{\textsc{NaQ}}$ for each support set sample $s_{\textsc{NaQ}}=(x_{\textsc{NaQ}}^{spt}, y_{\textsc{NaQ}}^{spt})\in S_\mathcal{T}$\footnote{Here, we use $S_\mathcal{T}$ to denote the support set generated by \textsc{NaQ}. For details, see `Support set generation' process in Section~\ref{naq:explain}.}
during the query generation process of \textsc{NaQ}. If $s$ and $q$ are class-level similar, i.e., the difference between their corresponding true labels $\tilde{y}_{\textsc{NaQ}}^{spt}, \tilde{y}_{\textsc{NaQ}}^{qry}$ are small enough, we would have $|\delta|<\xi$ for some small enough $\xi>0$ so that we can successfully train encoder $f_\theta$.

In summary, the above analysis explains that discovering a query that is class-level similar enough to a given support set sample is crucial for minimizing the training loss (i.e., the generalization error defined in Eq.~\ref{generr_meta}), which eventually yields a better $f_\theta$. 
\textbf{In this regard, \textsc{NaQ} works well within the episodic learning framework}, since \textsc{NaQ} generates class-level similar query nodes using node-node similarity defined based on the raw node feature (i.e., \textsc{NaQ-Feat}) and graph structural information (i.e., \textsc{NaQ-Diff}).

Further discussions on why class-level similarity is sufficient for unsupervised episode generation (Section~\ref{qualitative:analysis}) and an empirical result that our \textsc{NaQ} can find class-level similar queries (Section~\ref{analysis:empirical}) are provided in the Appendix.

\begin{table*}[ht!]
\centering
\caption{Overall averaged FSNC accuracy (\%) with 95\% confidence intervals on citation networks (Full ver. available at: Table~\ref{main_table:full:cfdb}, OOT: Out Of Time, which means that the training was not finished in 24 hours, OOM: Out Of Memory on NVIDIA RTX A6000)}
\vspace{1mm}
\label{main_table:cfdb}
\renewcommand{\arraystretch}{1.2}
\resizebox{0.95\textwidth}{!}{%
\begin{tabular}{c|ccccccc|ccccccc}
\hline
Dataset & \multicolumn{7}{c|}{\textbf{Cora-full}} & \multicolumn{7}{c}{\textbf{DBLP}} \\ \hline
Setting & \multicolumn{2}{c}{5 way} & \multicolumn{2}{c}{10 way} & \multicolumn{2}{c|}{20 way} & \multirow{2}{*}{\begin{tabular}[c]{@{}c@{}}Avg.\\ Rank\end{tabular}} & \multicolumn{2}{c}{5 way} & \multicolumn{2}{c}{10 way} & \multicolumn{2}{c|}{20 way} & \multirow{2}{*}{\begin{tabular}[c]{@{}c@{}}Avg.\\ Rank\end{tabular}} \\ \cline{1-7} \cline{9-14}
Baselines & 1 shot & 5 shot & 1 shot & 5 shot & 1 shot & \multicolumn{1}{c|}{5 shot} &  & 1 shot & 5 shot & 1 shot & 5 shot & 1 shot & \multicolumn{1}{c|}{5 shot} &  \\ \hline
MAML (Sup.) & 59.28±\tiny{1.21} & 70.30±\tiny{0.99} & 44.15±\tiny{0.81} & 57.59±\tiny{0.66} & 30.99±\tiny{0.43} & \multicolumn{1}{c|}{46.80±\tiny{0.38}} & 9.67 & 72.48±\tiny{1.22} & 80.30±\tiny{1.03} & 60.08±\tiny{0.90} & 69.85±\tiny{0.76} & 46.12±\tiny{0.53} & \multicolumn{1}{c|}{57.30±\tiny{0.48}} & 8.50 \\
ProtoNet (Sup.) & 58.61±\tiny{1.21} & 73.91±\tiny{0.93} & 44.54±\tiny{0.79} & 62.15±\tiny{0.64} & 32.10±\tiny{0.42} & \multicolumn{1}{c|}{50.87±\tiny{0.40}} & 7.67 & 73.80±\tiny{1.20} & 81.33±\tiny{1.00} & 61.88±\tiny{0.86} & 73.02±\tiny{0.74} & 48.70±\tiny{0.52} & \multicolumn{1}{c|}{62.42±\tiny{0.45}} & 4.33 \\
TENT (Sup.) & 61.30±\tiny{1.18} & 77.32±\tiny{0.81} & 47.30±\tiny{0.80} & 66.40±\tiny{0.62} & 36.40±\tiny{0.45} & \multicolumn{1}{c|}{55.77±\tiny{0.39}} & 4.50 & 74.01±\tiny{1.20} & \underline{82.54±\tiny{1.00}} & \underline{62.95±\tiny{0.85}} & \underline{73.26±\tiny{0.77}} & 49.67±\tiny{0.53} & \multicolumn{1}{c|}{61.87±\tiny{0.47}} & \underline{2.67} \\
G-Meta (Sup.) & 59.88±\tiny{1.26} & 75.36±\tiny{0.86} & 44.34±\tiny{0.80} & 59.59±\tiny{0.66} & 33.25±\tiny{0.42} & \multicolumn{1}{c|}{49.00±\tiny{0.39}} & 7.50 & \underline{74.64±\tiny{1.20}} & 79.96±\tiny{1.08} & 61.50±\tiny{0.88} & 70.33±\tiny{0.77} & 46.07±\tiny{0.52} & \multicolumn{1}{c|}{58.38±\tiny{0.47}} & 7.00 \\
GLITTER (Sup.) & 55.17±\tiny{1.18} & 69.33±\tiny{0.96} & 42.81±\tiny{0.81} & 52.76±\tiny{0.68} & 30.70±\tiny{0.41} & \multicolumn{1}{c|}{40.82±\tiny{0.41}} & 11.50 & 73.50±\tiny{1.25} & 75.90±\tiny{1.19} & OOT & OOT & OOM & \multicolumn{1}{c|}{OOM} & 9.50 \\
COSMIC (Sup.) & 62.24±\tiny{1.15} & 73.85±\tiny{0.83} & 47.85±\tiny{0.77} & 59.11±\tiny{0.60} & 42.25±\tiny{0.43} & \multicolumn{1}{c|}{47.28±\tiny{0.38}} & 6.33 & 72.34±\tiny{1.18} & 80.83±\tiny{1.03} & 59.21±\tiny{0.80} & 70.67±\tiny{0.71} & 49.52±\tiny{0.51} & \multicolumn{1}{c|}{59.01±\tiny{0.42}} & 7.50 \\ \hline
TLP-BGRL & 62.59±\tiny{1.13} & 78.80±\tiny{0.80} & 49.43±\tiny{0.79} & 67.18±\tiny{0.61} & 37.63±\tiny{0.44} & \multicolumn{1}{c|}{56.26±\tiny{0.39}} & 3.17 & 73.92±\tiny{1.19} & 82.42±\tiny{0.95} & 60.16±\tiny{0.87} & 72.13±\tiny{0.74} & 47.00±\tiny{0.53} & \multicolumn{1}{c|}{60.57±\tiny{0.45}} & 4.83 \\
TLP-SUGRL & 55.42±\tiny{1.08} & 76.01±\tiny{0.84} & 44.66±\tiny{0.74} & 63.69±\tiny{0.62} & 34.23±\tiny{0.41} & \multicolumn{1}{c|}{52.76±\tiny{0.40}} & 6.33 & 71.27±\tiny{1.15} & 81.36±\tiny{1.02} & 58.85±\tiny{0.81} & 71.02±\tiny{0.78} & 45.71±\tiny{0.49} & \multicolumn{1}{c|}{59.77±\tiny{0.45}} & 8.17 \\
TLP-AFGRL & 55.24±\tiny{1.02} & 75.92±\tiny{0.83} & 44.08±\tiny{0.70} & 64.42±\tiny{0.62} & 33.88±\tiny{0.41} & \multicolumn{1}{c|}{53.83±\tiny{0.39}} & 7.17 & 71.18±\tiny{1.17} & 82.03±\tiny{0.94} & 58.70±\tiny{0.86} & 71.14±\tiny{0.75} & 45.99±\tiny{0.53} & \multicolumn{1}{c|}{60.31±\tiny{0.45}} & 7.83 \\ \hline
VNT & 47.53±\tiny{1.14} & 69.94±\tiny{0.89} & 37.79±\tiny{0.69} & 57.71±\tiny{0.65} & 28.78±\tiny{0.40} & \multicolumn{1}{c|}{46.86±\tiny{0.40}} & 11.17 & 58.21±\tiny{1.16} & 76.25±\tiny{1.05} & 48.75±\tiny{0.81} & 66.37±\tiny{0.77} & 40.10±\tiny{0.49} & \multicolumn{1}{c|}{55.15±\tiny{0.46}} & 11.17 \\ \hline
\textbf{\textsc{NaQ-Feat}-Best (Ours)} & \textbf{66.30±\tiny{1.15}} & \textbf{80.09±\tiny{0.79}} & \textbf{52.23±\tiny{0.73}} & \underline{68.87±\tiny{0.60}} & \textbf{44.13±\tiny{0.47}} & \multicolumn{1}{c|}{\underline{60.94±\tiny{0.36}}} & \textbf{1.33} & 73.55±\tiny{1.16} & 82.36±\tiny{0.94} & 60.70±\tiny{0.87} & 72.36±\tiny{0.73} & \underline{50.42±\tiny{0.52}} & \multicolumn{1}{c|}{\textbf{64.90±\tiny{0.43}}} & 3.67 \\
\textbf{\textsc{NaQ-Diff}-Best (Ours)} & \underline{66.26±\tiny{1.15}} & \underline{80.07±\tiny{0.79}} & \underline{52.17±\tiny{0.74}} & \textbf{69.34±\tiny{0.63}} & \underline{44.12±\tiny{0.47}} & \multicolumn{1}{c|}{\textbf{60.97±\tiny{0.37}}} & \underline{1.67} & \textbf{76.58±\tiny{1.18}} & \textbf{82.86±\tiny{0.95}} & \textbf{64.31±\tiny{0.87}} & \textbf{74.06±\tiny{0.75}} & \textbf{51.62±\tiny{0.54}} & \multicolumn{1}{c|}{\underline{64.78±\tiny{0.44}}} & \textbf{1.17} \\ \hline
\end{tabular}%
}
\end{table*}

\section{Experiments}

\noindent \textbf{Evaluation Datasets. }
We use five benchmark datasets that are widely used in FSNC to comprehensively evaluate the performance of our unsupervised episode generation
method: 1) Two product networks (\textbf{Amazon-Clothing}, \textbf{Amazon-Electronics}~\cite{amazonnetwork}), 2) three citation networks (\textbf{Cora-Full}~\cite{corafull}, \textbf{DBLP}~\cite{dblp}) in addition to a large-scale dataset \textbf{ogbn-arxiv}~\cite{hu2020ogb}.
Detailed explanations of the datasets and their statistics are provided in Section~\ref{detail:dataset} in the Appendix.

\noindent \textbf{Baselines. }
We use six graph meta-learning models as baselines, {i.e.}, \textbf{MAML}~\cite{maml}, \textbf{ProtoNet}~\cite{protonet}, \textbf{G-Meta}~\cite{gmeta}, \textbf{TENT}~\cite{tent}, \textbf{GLITTER}~\cite{glitter}, and \textbf{COSMIC}~\cite{cosmic} to evaluate the performance of our proposed unsupervised episode generation methods, {i.e.}, \textsc{NaQ-Feat} and \textsc{NaQ-Diff}. In addition, three recent GCL baselines, {i.e.}, \textbf{BGRL}~\cite{bgrl}, \textbf{SUGRL}~\cite{sugrl} and \textbf{AFGRL}~\cite{afgrl}, are included as they have shown remarkable performance on the FSNC task without using labels~\cite{tlp}. 
Lastly, we compare with \textbf{VNT}~\cite{vnt} that uses a pretrained graph transformer without labels and fine-tunes injected soft prompts to solve downstream FSNC task. 
For both \textsc{NaQ-Feat} and \textsc{NaQ-Diff}, we sampled $Q=10$ queries for each support set sample to generate the training episodes. 
Details on compared baselines and their experimental settings are presented in Section~\ref{detail:baseline} in the Appendix.

\noindent \textbf{Evaluation. }
For each dataset except for Amazon-Clothing, we evaluate the performance of the models in 5/10/20-way, 1/5-shot settings, i.e., six settings in total. For Amazon-Clothing, as the validation set contains 17 classes, evaluations on 20-way cannot be conducted. Instead, the evaluation is done in 5/10-way 1/5-shot settings, i.e., four settings in total. In the validation and testing phases, we sampled 50 validation tasks and 500 testing tasks for all settings with 8 queries each.
{For all the baselines, validation/testing tasks are fixed, and we use linear probing on frozen features to solve each downstream task except for GLITTER and VNT as they use different strategies for solving downstream tasks.} We report average accuracy and 95\% confidence interval over sampled testing tasks.

\subsection{Overall Performance Analysis} \label{overall:analysis}
The overall results on five datasets are presented in Table~\ref{main_table:acae},~\ref{main_table:cfdb}, {and~\ref{table:arxiv}.}
Note that since \textsc{NaQ} is \textit{model-agnostic}, we apply \textsc{NaQ} with all the supervised graph meta-learning models contained in our baselines, and report the best performance among them.
We have the following observations.

\begin{figure*} [t!]
    \centering
    \includegraphics[width=0.99\textwidth]{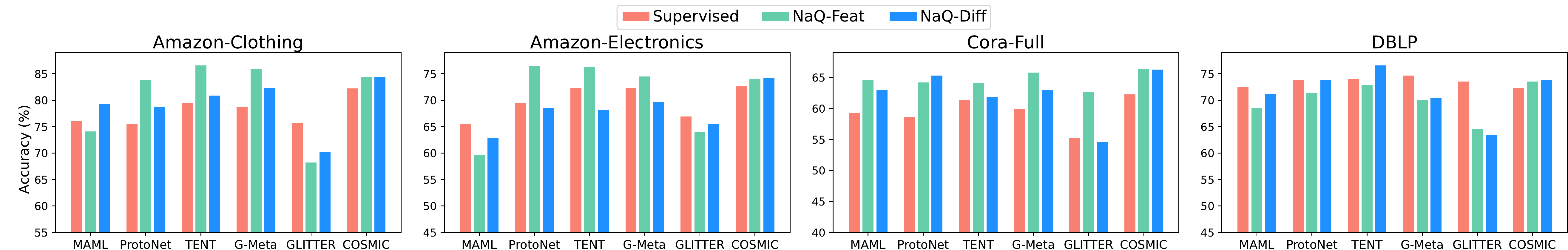}
    \vspace{-2ex}
    \caption{Result of applying \textsc{NaQ-Feat} and \textsc{NaQ-Diff} to existing graph meta-learning models (5-way 1-shot).}
    \label{naq_model_agnostic}
\end{figure*}

\begin{figure*} [t!]
    \centering
    \includegraphics[width=0.99\textwidth]{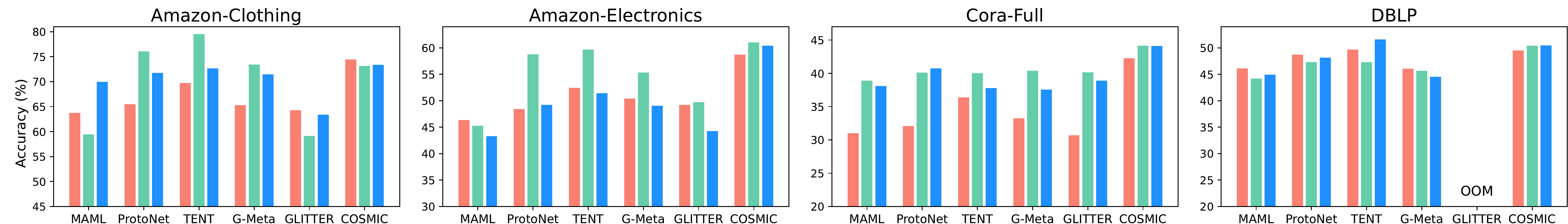}
    \vspace{-2ex}
    \caption{Result of applying \textsc{NaQ-Feat} and \textsc{NaQ-Diff} to existing graph meta-learning models in higher way settings (Amazon-Clothing: 10-way 1-shot, Others: 20-way 1-shot).}
    \label{naq_model_agnostic_higher_way}
    \vspace{-1ex}
\end{figure*}

First, our proposed methods outperform the existing supervised baselines. 
We attribute this to the episode generation strategy of \textsc{NaQ} that allows the model to extensively utilize all nodes in the graph without reliance on node labels.
It is worth noting that for each training episode while other supervised methods use 5-shot support sets, our methods use 1-shot support sets to ensure that the support set nodes are as much distinguishable from one another as possible. Hence, we expect that our methods can be further improved if we develop methods to generate additional support set samples that would make each support set even more distinguishable from one another, which we leave as future work.

Second, our proposed methods outperform methods utilizing the pre-trained encoder in an unsupervised manner ({i.e.}, GCL methods and VNT). Unlike these methods, by applying the episodic learning framework for model training, our methods can capture information about the downstream task `format' during the model training, leading to generally better performance in the FSNC task.

Third, \textsc{NaQ-Diff} outperforms \textsc{NaQ-Feat} in citation networks (Table~\ref{main_table:cfdb}). This result verifies our motivation for presenting \textsc{NaQ-Diff} in Section~\ref{naq_diff:rise}, which was to capture the structural information instead of the raw node features in domains where the structural information is more beneficial.
On the other hand, \textsc{NaQ-Feat} outperforms \textsc{NaQ-Diff} in product networks. This is because products in `also-viewed' or `bought-together' relationships are {not always} similar or related in case of product networks~\cite{hsd_seq_recomm}, implying that discovering query sets based on `raw-feature' similarity is more beneficial.

Lastly, \textsc{NaQ} outperforms other baselines on the ogbn-arxiv dataset, which is a large-scale dataset (Table~\ref{table:arxiv}). It is worth noting that the performances of two variants of \textsc{NaQ} are at the best and second best in a more challenging one-shot setting. One interesting observation is that \textsc{NaQ-Feat} outperforms \textsc{NaQ-Diff}, even though ogbn-arxiv is a citation network. We attribute this to the fact that the raw node features of ogbn-arxiv are `embeddings' extracted from the skip-gram model. This implies that \textit{high-quality node feature enables \textsc{NaQ-Feat} to find high-quality queries, which leads to a better FSNC performance of \textsc{NaQ-Feat}}. 

In summary, \textsc{NaQ} resolves the label-scarcity problem of supervised graph meta-learning methods and achieve performance enhancement on FSNC tasks by providing training episodes that contain both the information of all nodes in the graph, and the information of the downstream task format to the model.

\subsection{Model-agnostic Property of \textsc{NaQ}} \label{naq:withsup}
In this section, we verify that \textsc{NaQ} can be applied to any existing graph meta-learning models while not sacrificing much of their performance.

In Figure~\ref{naq_model_agnostic} and~\ref{naq_model_agnostic_higher_way}, 
we observe that our methods retained or even improved the performance of existing graph meta-learning methods across various few-shot settings with only a few exceptions. 
Particularly, in higher way settings shown in Figure~\ref{naq_model_agnostic_higher_way}, which are more challenging, \textsc{NaQ} generally outperforms supervised methods.
Therefore, we argue that our methods allow existing graph meta-learning models to be trained to generate more generalizable embeddings without any use of label information thanks to the full utilization of all nodes in a graph.

Lastly, it is important to note again that the performances of the supervised models reported in our experiments are only achievable \textit{when they have access to all labeled samples of entire base classes and the given labeled samples in the base classes are clean.} On the other hand, when there is only a limited amount of labeled samples within a limited number of base classes (Figure~\ref{intro:sup_label}) or there is inherent label noise in the base classes (Figure~\ref{intro:label_noise}), the performance of supervised models severely drops, while our proposed unsupervised methods would not be affected at all. Furthermore, as will be demonstrated in Section~\ref{hyparam:analysis}, the performance of \textsc{NaQ} can be improved by adjusting the number of queries.

\begin{table}
\vspace{-2ex}
\centering
\caption{Overall averaged FSNC accuracy (\%) with 95\% confidence intervals on ogbn-arxiv (\textsc{NaQ} base-model: ProtoNet, OOM: Out Of Memory on NVIDIA RTX A6000)}
\vspace{1mm}
\label{table:arxiv}
\renewcommand{\arraystretch}{1.2}
\resizebox{0.99\columnwidth}{!}{%
\begin{tabular}{c|cccc}
\hline
Dataset         & \multicolumn{4}{c}{\textbf{ogbn-arxiv}}                                                                \\ \hline
Setting         & \multicolumn{2}{c}{5 way}                     & \multicolumn{2}{c}{10 way}                    \\ \hline
Baselines       & 1 shot             & 5 shot                   & 1 shot             & 5 shot                   \\ \hline
MAML (Sup.)     & 40.61±\tiny{0.89} & 58.75±\tiny{0.89}       & 27.32±\tiny{0.55} & 43.87±\tiny{0.56}       \\
ProtoNet (Sup.) & 43.34±\tiny{1.01} & 58.30±\tiny{0.95}       & 28.17±\tiny{0.60} & 46.11±\tiny{0.60}       \\
TENT (Sup.)     & 48.06±\tiny{0.97} & 63.45±\tiny{0.88}       & 33.85±\tiny{0.65} & 48.14±\tiny{0.59}       \\
G-Meta (Sup.)   & 41.06±\tiny{0.87} & 59.43±\tiny{0.87}       & 27.20±\tiny{0.53} & 45.04±\tiny{0.53}       \\
GLITTER (Sup.)  & 35.64±\tiny{0.97} & 34.51±\tiny{0.85}       & 20.95±\tiny{0.50} & 21.84±\tiny{0.47}       \\
COSMIC (Sup.)   & 50.32±\tiny{0.95} & 63.54±\tiny{0.80}       & 38.41±\tiny{0.62} & 49.31±\tiny{0.51}       \\ \hline
TLP-BGRL        & 49.88±\tiny{1.01} & \underline{69.10±\tiny{0.82}} & 36.40±\tiny{0.62} & \underline{56.15±\tiny{0.54}} \\
TLP-SUGRL       & 49.25±\tiny{0.97} & 62.15±\tiny{0.92}       & 32.87±\tiny{0.61} & 45.76±\tiny{0.60}       \\
TLP-AFGRL       & OOM                & OOM                      & OOM                & OOM                      \\ \hline
VNT             & OOM                & OOM                      & OOM                & OOM                      \\ \hline
\textbf{\textsc{NaQ-Feat} (Ours)} & \textbf{54.09±\tiny{1.03}} & \textbf{69.94±\tiny{0.84}} & \textbf{41.61±\tiny{0.68}} & \textbf{58.18±\tiny{0.59}} \\
\textbf{\textsc{NaQ-Diff} (Ours)} & \underline{51.45±\tiny{1.04}}    & 66.73±\tiny{0.89}          & \underline{39.27±\tiny{0.67}}    & 55.93±\tiny{0.56}          \\ \hline
\end{tabular}%
}
\end{table}

\subsection{Regarding the Class Imbalance} \label{exp:class_imbal}
In this section, we visualize t-SNE~\cite{tsne} embeddings of nodes that belong to the top-10 tail classes.
Doing so can further justify our motivation for using unsupervised graph meta-learning on FSNC problems rather than using GCL methods.

\begin{figure} [t]
    \centering
        \subfigure[Amazon-Electronics]{
        \includegraphics[width=0.99\linewidth]{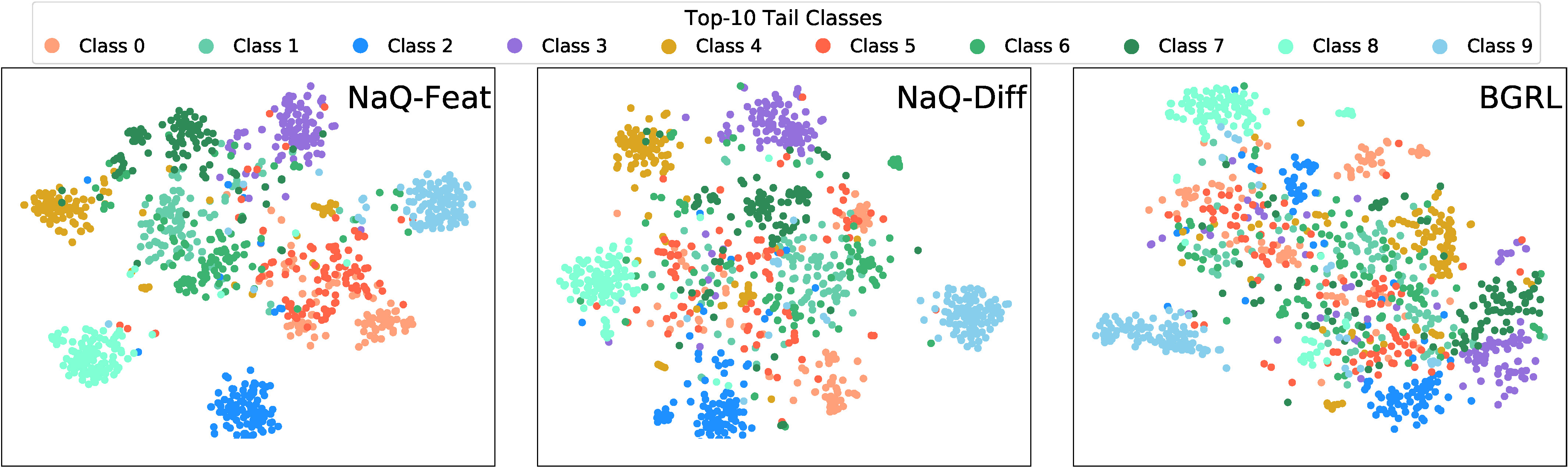}
        \label{class_imbal_tsne:ae}
        }

        \subfigure[Cora-Full]{
        \includegraphics[width=0.99\linewidth]{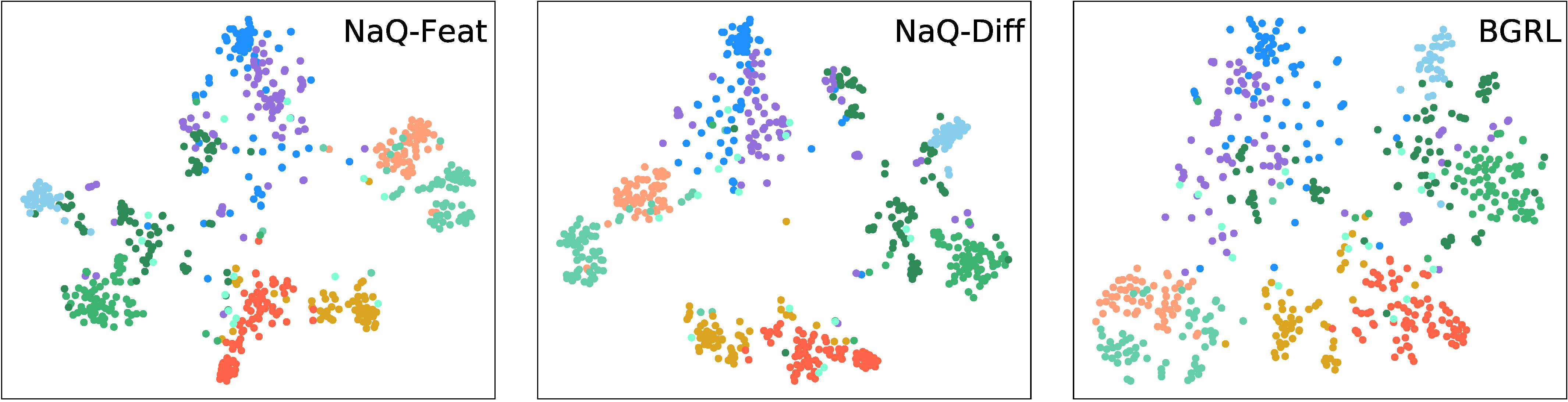}
        \label{class_imbal_tsne:cf}
        \vspace{-3mm}
        }
    \vspace{-4mm}
    \caption{The t-SNE plot of tail-class embeddings (base-model: ProtoNet, \textsc{NaQ}: trained with 5-way 1-shot training episodes)}
    \label{class_imbal_tsne}
    \vspace{-3ex}
\end{figure}

As shown in Figure~\ref{class_imbal_tsne}, we can observe that our \textsc{NaQ} can learn clearly separable embeddings for tail-class nodes than GCL method BGRL. This result further supports our claim that GCL methods have difficulty in learning embeddings of nodes from minority classes. Therefore, we can verify that additional downstream task `format' information provided by episodic learning is beneficial for learning tail-class nodes when solving the FSNC problem. Further discussions on why \textsc{NaQ} can attain robustness against the class imbalance (Section~\ref{imbal:discuss}) and additional results on various dataset biases, such as structure or feature noise (Section~\ref{appendix:bias:result}), are presented in the Appendix.

\subsection{Hyperparameter Sensitivity Analysis} \label{hyparam:analysis}
\vspace{-1ex}
So far, the experiments have been conducted with a fixed number of queries, $Q=10$. In this section, we investigate the effect of the number of queries on the performance of \textsc{NaQ}.
To thoroughly explore the effect of the number of queries on \textsc{NaQ}, we check the performance of \textsc{NaQ} with ProtoNet by changing the number of queries $Q\in$\{1, 3, 5, 7, 10, 13, 15, 17, 20, 30, 40, 100\}. 
We have the following observations from Figure~\ref{hyparam:study:naq}: 
\textbf{(1)} In Amazon-Clothing, since both \textsc{NaQ-Feat} and \textsc{NaQ-Diff} can discover highly class-level similar queries (Figure~\ref{analysis:classsim}), they exhibit an increasing tendency in performance as $Q$ increases.
\textbf{(2)} In the case of Amazon-Electronics, \textsc{NaQ-Feat} shows a similar tendency as in Amazon-Clothing due to the same reason, while there is a slight performance drop when $Q=100$. In contrast, \textsc{NaQ-Diff} shows clearly decreasing performance after $Q=5$, as its queries have relatively low class-level similarity (Figure~\ref{analysis:classsim}). From the results above, we can conclude that sampling a proper number of queries $Q$ during the episode generation phase is essential. Otherwise, a significant level of label noise in the generated episode might hinder the model training.
{\textbf{(3)} In the DBLP dataset, \textsc{NaQ-Feat} shows a nearly consistent performance tendency, while the performance of \textsc{NaQ-Diff} can be enhanced by increasing the number of queries for training. This is because \textsc{NaQ-Diff} can sample more class-level similar queries than \textsc{NaQ-Feat} (Figure~\ref{analysis:classsim}). From this observation, we again validate the motivation of utilizing structural neighbors as queries in such datasets (Section~\ref{naq_diff:rise}).}

\begin{figure} [t]
    \centering
        \subfigure[\textsc{NaQ-Feat}]{
        \includegraphics[width=0.45\columnwidth]{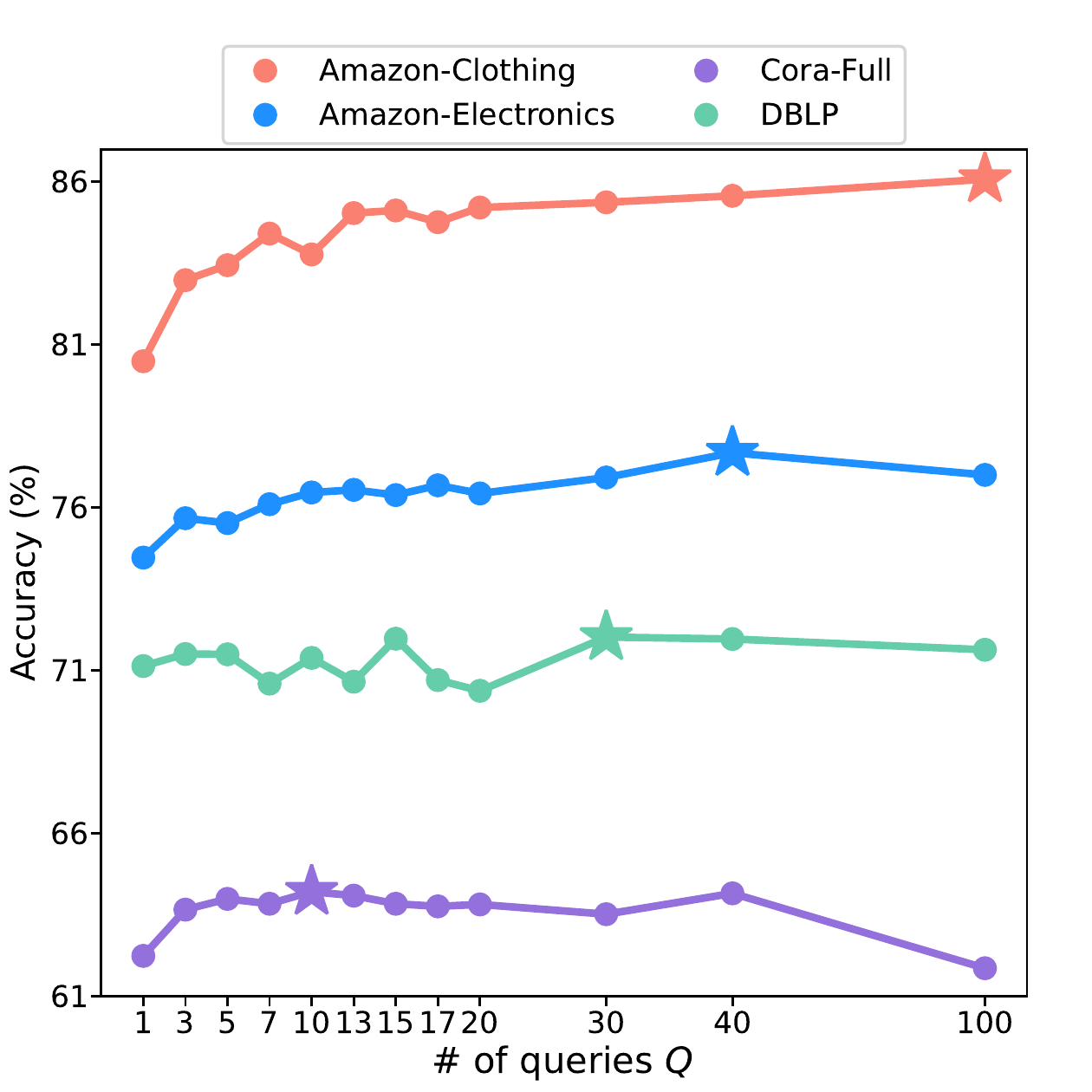}
        \label{hyparam:naq:feat}
        \vspace{-3mm}
        }
        \hspace{1mm}
        \subfigure[\textsc{NaQ-Diff}]{
        \includegraphics[width=0.45\columnwidth]{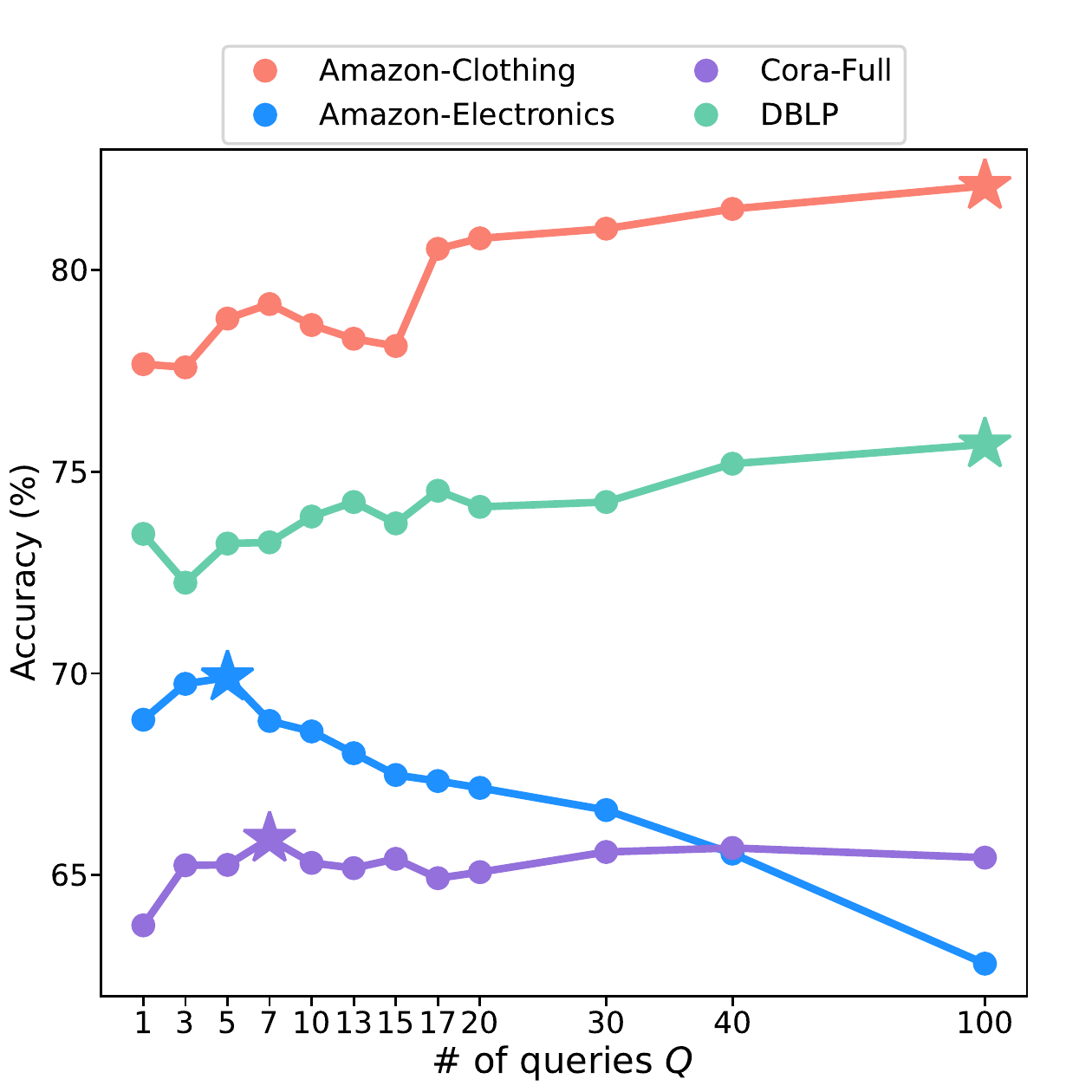}
        \label{hyparam:naq:diff}
        \vspace{-3mm}
        }
    \vspace{-3mm}
    \caption{Effect of the number of queries \textsc{NaQ} (5-way 1-shot, base-model: ProtoNet, star marker: maximal point)}
    \label{hyparam:study:naq}
    \vspace{-3ex}
\end{figure}

\section{Related Work}
\subsection{Few-Shot Node Classification (FSNC)}
Few-shot learning~\cite{vinyals2016matching, maml, protonet} aims to classify unseen target classes with only a few labeled samples based on the meta-knowledge obtained from training on abundant samples from base classes.

\noindent\textbf{Graph Meta-learning. } There have been various studies to solve FSNC in graph-structured data.
Meta-GNN~\cite{metagnn} addresses the problem by directly applying MAML~\cite{maml} on GNN, and GPN~\cite{gpn} uses ProtoNet~\cite{protonet} architecture with adjusted prototype calculation by considering node importance.
G-Meta~\cite{gmeta} utilizes subgraph-level embeddings of nodes inside training episodes based on both ProtoNet and MAML frameworks to enable scalable and inductive graph meta-learning.
TENT~\cite{tent} tries to reduce variances within training episodes through node-level, class-level, and task-level adaptations.
Meta-GPS~\cite{meta-gps} utilizes various components of network encoder, prototype-based parameter initialization, and $S^2$ (scaling \& shifting) transformation to solve FSNC tasks even on heterophilic graphs. GLITTER~\cite{glitter} claims that the given entire graph structure is redundant for learning node embeddings within the meta-task so that it tries to learn task-specific structure for each meta-task.
COSMIC~\cite{cosmic} applies a contrastive learning scheme on meta-learning to obtain the intra-class generalizability with hard (unseen) node classes generated by similarity-sensitive mix-up to achieve high inter-class generalizability.

\noindent\textbf{Graph Meta-learning for Label-scarcity Problem. } There were a few studies aiming to alleviate the label-scarcity problem of graph meta-learning methods.
TEG~\cite{teg} utilizes equivariant neural networks to capture task-patterns shared among training episodes regardless of node labels, enabling the learning of highly transferable task-adaptation strategies even with a limited number of base classes and labeled nodes.
Meanwhile, X-FNC~\cite{x-fnc} obtains pseudo-labeled nodes via label propagation based on Poisson Learning, and optimizes the model based on information bottleneck to discard irrelevant information within the augmented support set.
Although these methods extract useful meta-knowledge based on training episodes (i.e., TEG) or from pseudo-labeled nodes (i.e., X-FNC), they still highly depend on a few labeled nodes during the model training, and thus still fall short of utilizing the information of all nodes in the graph. As a result, their FSNC performance degrades as the number of labeled nodes and base classes decreases~\cite{x-fnc, teg}.

\noindent\textbf{Unsupervised FSNC. } As existing graph meta-learning methods suffer from the label-scarcity problem, there were several studies to handle the FSNC problem in an unsupervised manner. TLP~\cite{tlp} utilizes GCL methods to solve FSNC, and it has shown superior FSNC performance than graph meta-learning methods without labels.
VNT~\cite{vnt} applies graph transformer on FSNC and solves downstream FSNC task by only fine-tuning `virtual' nodes injected as soft prompts and the classifier with given a few-labeled samples in the downstream task.
Most recently, \cite{cola} analyse advantages of applying GCL on FSNC over graph meta-learning in two aspects: 1) utilization of graph augmentation, and 2) explicit usage of all nodes in a graph. Base on this analysis, they present a GCL-based method named COLA that aims to combine GCL and meta-learning by constructing meta-tasks without labels \textit{during} the training phase, which is computationally costly. Although it shares some similarities with our method \textsc{NaQ}, COLA focuses on GCL-based model while our \textsc{NaQ} focuses on enabling unsupervised graph meta-learning.

\subsection{Unsupervised Meta-learning}
In computer vision, several unsupervised meta-learning methods exist that attempt to address the limitations of requiring abundant labels for constructing training episodes. More precisely, UMTRA~\cite{umtra} and AAL~\cite{aal} are similar methods, making queries via image augmentation on randomly sampled support set samples. 
In addition, AAL focuses on task generation, while UMTRA is mainly applied to MAML. On the other hand, CACTUs~\cite{cactus} aims to make episodes based on pseudo-labels obtained from cluster assignments, which come from features pre-trained in an unsupervised fashion. LASIUM~\cite{lasium} generates synthetic training episodes that can be combined with existing models, such as MAML and ProtoNet, with generative models. Moreover, Meta-GMVAE~\cite{metagmvae} uses VAE~\cite{vae} with Gaussian mixture priors to solve the few-shot learning problem.

\section{Limitations \& Future Work}
Although \textsc{NaQ} has proven its effectiveness for Few-Shot Node Classification (FSNC), it is crucial to acknowledge its limitations, presented below, to stimulate future work.

\subsection{Computational Issue of \textsc{NaQ-Diff}}
Due to some technical issues regarding sparse matrix multiplication, we cannot even calculate the truncated approximation of the graph Diffusion for the dataset, which has many edges (e.g., ogbn-products).
This problem hinders the applicability of \textsc{NaQ-Diff} to large real-world datasets.
Hence, it will be promising to devise unsupervised episode generation methods that can fully leverage the structural information of graphs while reducing computational costs.

\subsection{Problem of Na\"{i}ve Support Set Generation.}
Since \textsc{NaQ} depends on na\"{i}ve random sampling for support set generation, there is a possibility that nodes having the same label can be assigned to a distinct support set, which is an undesirable case. Although we sample 1-shot support sets to avoid the above problem, developing a more sophisticated support set generation method that mitigates the problem mentioned above and generates a $K$-shot ($K>1$) support set will be valuable future work.

\section{Conclusion}
In this study, we proposed \textsc{NaQ}, a novel unsupervised episode generation algorithm that enables unsupervised graph meta-learning.
\textsc{NaQ} generates 1) support sets by random sampling from the entire graph, and 2) query sets by utilizing feature-level similar nodes ({i.e.,} \textsc{NaQ-Feat}) or structurally similar neighbors from graph diffusion ({i.e.,} \textsc{NaQ-Diff}). As \textsc{NaQ} generates training episodes out of all nodes in the graph without any label information, it can address the label-scarcity problem of supervised graph meta-learning models. Moreover, generated episodes from \textsc{NaQ} can be used for training any existing graph meta-learning models almost without modifications and even boost their performance on the FSNC task. Extensive experimental studies on various downstream task settings demonstrate the superiority and potential of \textsc{NaQ}.

\clearpage
\section*{Acknowledgment}
This work was supported by the National Research Foundation of Korea(NRF) grant funded by the Korea government(MSIT) (RS-2024-00335098), and supported by National Research Foundation of Korea(NRF) funded by Ministry of Science and ICT (NRF-2022M3J6A1063021).

\section*{Impact Statement}
This paper proposes to advance the field of unsupervised Machine Learning on graph-structured data like social networks. Although our research might have many potential ethical/societal impacts during its application, we believe no specific points should be emphasized here.


\bibliography{icml_naq}
\bibliographystyle{icml2024}

\newpage
\appendix
\onecolumn
\section{Appendix}

\subsection{Regarding `Class-level Similarity'} \label{why:class:sim}
\subsubsection{Why is `class-level similarity' sufficient?} \label{qualitative:analysis}
\looseness=-1
In Section~\ref{theo:insight}, we justified our `similarity' condition presented in Section~\ref{motivation} in terms of \textit{`class-level similarity'}.
In this section, we provide an explanation on why considering the class-level similarity instead of the exact same class condition, which is in fact impossible because the class information is not given, is sufficient for the query generation process in \textsc{NaQ}, {and further justify why our method outperforms supervised meta-learning methods.}

Overall, we conjecture that training a model via episodic learning with episodes generated from \textsc{NaQ} can be done successfully not only because our methods enable the utilization of all nodes in a graph, but also because our methods generate sufficiently informative episodes that enable the model to learn the downstream task format.
When we take a closer look at the training process of an episodic learning framework, the model only needs to classify a small number ($N$-way) of classes in a single episode unlike the conventional training scheme requiring the model to classify total $|C|$ classes in a graph.
For this reason, we do not have to strive for finding queries whose labels are exactly the same as their corresponding support set sample as in ordinary supervised episode generation.
Therefore, finding class-level similar queries is sufficient for generating informative training episodes.

Moreover, if we can generate training episodes that have queries similar enough to the corresponding support set sample while being dissimilar to the remaining $N-1$ support set samples, we further conjecture that the episodes utilizing class-level similar queries from \textsc{NaQ} is even more beneficial than episodes generated in the ordinary supervised manner.
This is because the episodes generated by \textsc{NaQ} provide helpful information from different but similar classes while episodes generated in the supervised manner merely provide the information within the same classes as support set sample.
To further demonstrate that \textsc{NaQ} has the ability to discover such class-level similar queries, empirical analysis is provided in Section~\ref{analysis:empirical}.

\begin{figure} [h!]
  \centering
  \includegraphics[width=0.55\columnwidth]{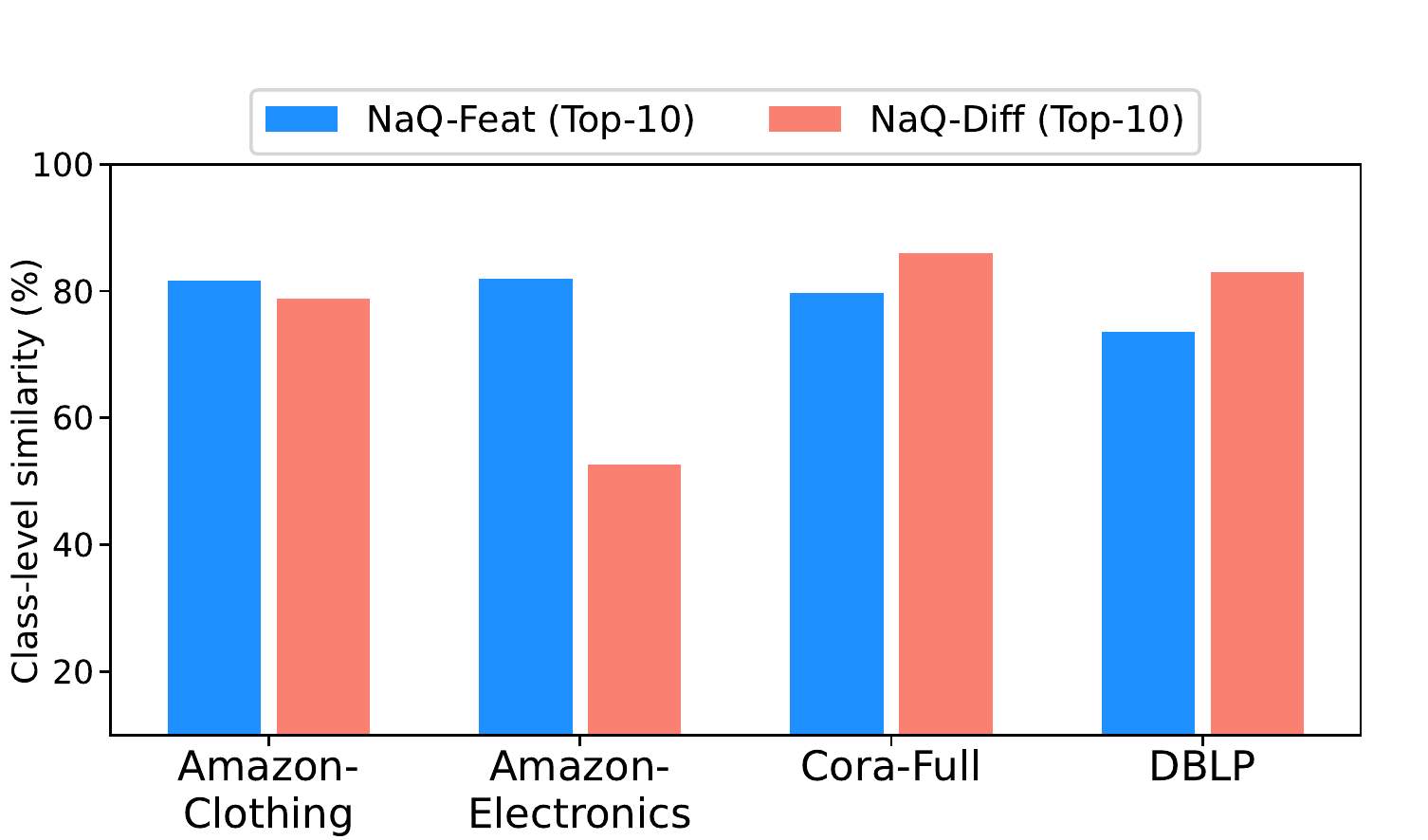}
  \vspace{-3mm}
  \caption{Averaged class-level similarity between each node and top-10 similar nodes found via \textsc{NaQ-Feat} and \textsc{NaQ-Diff}}
  \label{analysis:classsim}
\end{figure}

\subsubsection{\textsc{NaQ} discovers `class-level similar' queries } \label{analysis:empirical}
In this section, we provide an empirical evidence that \textsc{NaQ} can find class-level similar neighbors as queries for each support set sample (Figure~\ref{analysis:classsim}), and we further analyze the experimental results of our methods based on that evidence.

To verify that queries found by \textsc{NaQ} are class-level similar, we measure the averaged class-level similarity between a node and its top-10 similar nodes found by \textsc{NaQ-Feat} (raw feature similarity) and \textsc{NaQ-Diff} (graph diffusion) in all four datasets. The class-level similarity between two nodes is computed based on the similarity between their class centroids, where the centroid of class $c$ is computed by $\mathbf{a}_c=\mathrm{MEAN}(\sum{x_i\cdot\mathbb{I} \{y_i=c \})}$ with $x_i$ denoting the raw feature of node $i$ and $y_i$ denoting the label of node $i$.
The results are presented in Figure~\ref{analysis:classsim}. In most cases, similar nodes found by \textsc{NaQ-Feat} and \textsc{NaQ-Diff} exhibit a high-level ($\sim$80\%) average class-level similarity. This result shows that \textsc{NaQ-Feat} and \textsc{NaQ-Diff} can discover enough class-level similar nodes as queries for each support set sample. 

In addition, we can further justify our arguments in Section~\ref{naq_diff:rise} and the experimental results in Section~\ref{overall:analysis} based on these results. 
First, in Figure~\ref{analysis:classsim}, we observe that we can sample more class-level similar queries by \textsc{NaQ-Diff} than \textsc{NaQ-Feat} in citation networks ({i.e.,} Cora-Full and DBLP), implying that considering graph structural information can be more beneficial in citation networks for the reason described in Section~\ref{naq_diff:rise}. 
Second, since \textsc{NaQ-Diff} can discover class-level similar queries in the DBLP dataset, it shows superior performance in the DBLP dataset even though DBLP has a low homophily ratio. Therefore, we emphasize again that discovering class-level similar queries is essential in generating informative episodes.
Third, we observe that the variant of \textsc{NaQ} with higher class-level similarity always performs better in the downstream FSNC task, implying that making queries class-level similar to corresponding support set samples is directly related to the performance of \textsc{NaQ}. 

In summary, we quantitatively demonstrated that \textsc{NaQ} indeed discovers class-level similar nodes without using label information, and showed that the experimental results for our methods align well with our motivation regarding the support-query similarity, presented in Section~\ref{motivation} and justified in Section~\ref{theo:insight}.

\subsection{Regarding the Inherent Bias in Graphs} \label{appendix:bias}
Although in the main paper, we mentioned about the vulnerability of existing GCL methods to class imbalance, there exist other inherent bias that may exist in graphs, i.e., structure noise and feature noise. In this section, we provide further discussions on class imbalance (Section~\ref{imbal:discuss}) followed by additional results under structure and feature noise (Section~\ref{appendix:bias:result}).

\subsubsection{Further Discussion on Class Imbalance} \label{imbal:discuss}
In this section, we discuss in detail why existing graph meta-learning methods and our \textsc{NaQ} retains robustness against class imbalance in a graph. Even though we can conclude that task format information learned by episodic learning framework makes the model to be robust against the class imbalance from various experimental results (See Figure~\ref{intro:class_imbal} and Section~\ref{exp:class_imbal}), here we delve deeper into which elements of the (ordinary) supervised or unsupervised episode generation (\textsc{NaQ}) contribute to the robustness against the class imbalance. In addition, we present empirical analysis to further support our claim that episodic learning is beneficial to attain robustness against the class imbalance.

\textbf{Supervised Graph Meta-learning. } In the training episode generation step of the ordinary supervised meta-learning methods, they first sample $N$-way classes in base classes $C_b$, then sample $K$-shot support set samples and $Q$ queries within each of sampled classes. 
As a result, all classes \textit{in base classes} are treated equally regardless of the number of samples they contain. Therefore, with an aid of the task format information obtained via episodic learning, supervised graph meta-learning can be robust to class imbalance in a graph.

\begin{table}[h!]
\centering
\caption{Class-level similarity between each node from Top-$p$\% tail classes in the graph and top-10 similar nodes found via \textsc{NaQ-Feat} and \textsc{NaQ-Diff} (Results of 100\%: reported in Figure~\ref{analysis:classsim})}
\vspace{1mm}
\label{imbal:discuss_table}
\renewcommand{\arraystretch}{1.25}
\resizebox{0.9\columnwidth}{!}{%
\begin{tabular}{c|cc|cc|cc|cc}
\hline
Datasets &
  \multicolumn{2}{c|}{\textbf{Amazon-Clothing}} &
  \multicolumn{2}{c|}{\textbf{Amazon-Electronics}} &
  \multicolumn{2}{c|}{\textbf{Cora-Full}} &
  \multicolumn{2}{c}{\textbf{DBLP}} \\ \hline
\begin{tabular}[c]{@{}c@{}}top-$p$\%\\ tail classes\end{tabular} &
  \textsc{NaQ-Feat} &
  \textsc{NaQ-Diff} &
  \textsc{NaQ-Feat} &
  \textsc{NaQ-Diff} &
  \textsc{NaQ-Feat} &
  \textsc{NaQ-Diff} &
  \textsc{NaQ-Feat} &
  \textsc{NaQ-Diff} \\ \hline
10\%    & $\sim$78.7\% & $\sim$75.2\% & $\sim$72.3\% & $\sim$48.2\% & $\sim$69.7\% & $\sim$77.9\% & $\sim$66.6\% & $\sim$75.1\% \\
20\%    & $\sim$81.3\% & $\sim$78.2\% & $\sim$74.1\% & $\sim$51.6\% & $\sim$70.7\% & $\sim$77.6\% & $\sim$68.3\% & $\sim$78.0\% \\
50\%    & $\sim$81.7\% & $\sim$80.7\% & $\sim$77.8\% & $\sim$53.0\% & $\sim$74.6\% & $\sim$81.8\% & $\sim$70.4\% & $\sim$80.9\% \\
80\%    & $\sim$80.8\% & $\sim$79.0\% & $\sim$78.9\% & $\sim$52.5\% & $\sim$77.8\% & $\sim$84.6\% & $\sim$71.9\% & $\sim$82.1\% \\ \hline
100\%   & $\sim$81.6\% & $\sim$78.8\% & $\sim$81.9\% & $\sim$52.7\% & $\sim$79.8\% & $\sim$86.0\% & $\sim$73.5\% & $\sim$83.0\% \\ \hline
\end{tabular}%
}
\end{table}

\textbf{Unsupervised Graph Meta-learning with \textsc{NaQ}. } Since the class label information is not given to \textsc{NaQ}, addressing class imbalance is not trivial as in the supervised case described above.
Instead, \textsc{NaQ} samples `class-level similar' queries to the support set nodes from tail classes, which can help learning tail-class embeddings.
To demonstrate that \textsc{NaQ} still finds `class-level similar' queries to the tail-class nodes, we measured the averaged class-level similarity between node of the top-$p$\% tail classes and top-10 similar nodes found by \textsc{NaQ}. Results can be found in Table~\ref{imbal:discuss_table}. We observe that \textsc{NaQ} still finds class-level similar enough queries even for the nodes from tail classes, especially in the dataset in which each variant of \textsc{NaQ} outperforms (i.e., \textsc{NaQ-Feat} for product networks (Amazon-Clothing/Electronics), and \textsc{NaQ-Diff} for citation networks (Cora-Full, DBLP)). For top-10\% tail classes, queries found by \textsc{NaQ-Feat} exhibit 78.67\% / 72.29\% class-level similarity in Amazon-Clothing / Amazon-Electronics, and queries found by \textsc{NaQ-Diff} exhibit 77.89\% / 75.05\% class-level similarity in Cora-Full / DBLP. Therefore, we can conclude that `class-level similar' queries found by \textsc{NaQ} are beneficial for learning tail-class embeddings from the results of Table~\ref{imbal:discuss_table} and Section~\ref{exp:class_imbal}.

\textbf{Role of the Episodic Learning Framework. } To empirically examine whether downstream task format information provided by episodic learning helps attain robustness against the class imbalance in the graph or not, we observed the change in the quality of t-SNE embeddings of the top-10 tail-class nodes produced by \textsc{NaQ-Diff} when $N$-way becomes larger (i.e., $N=5\rightarrow20$, more challenging training setting) in the Amazon-Electronics dataset.

\begin{figure} [h!]
    \centering
    \includegraphics[width=0.6\columnwidth]{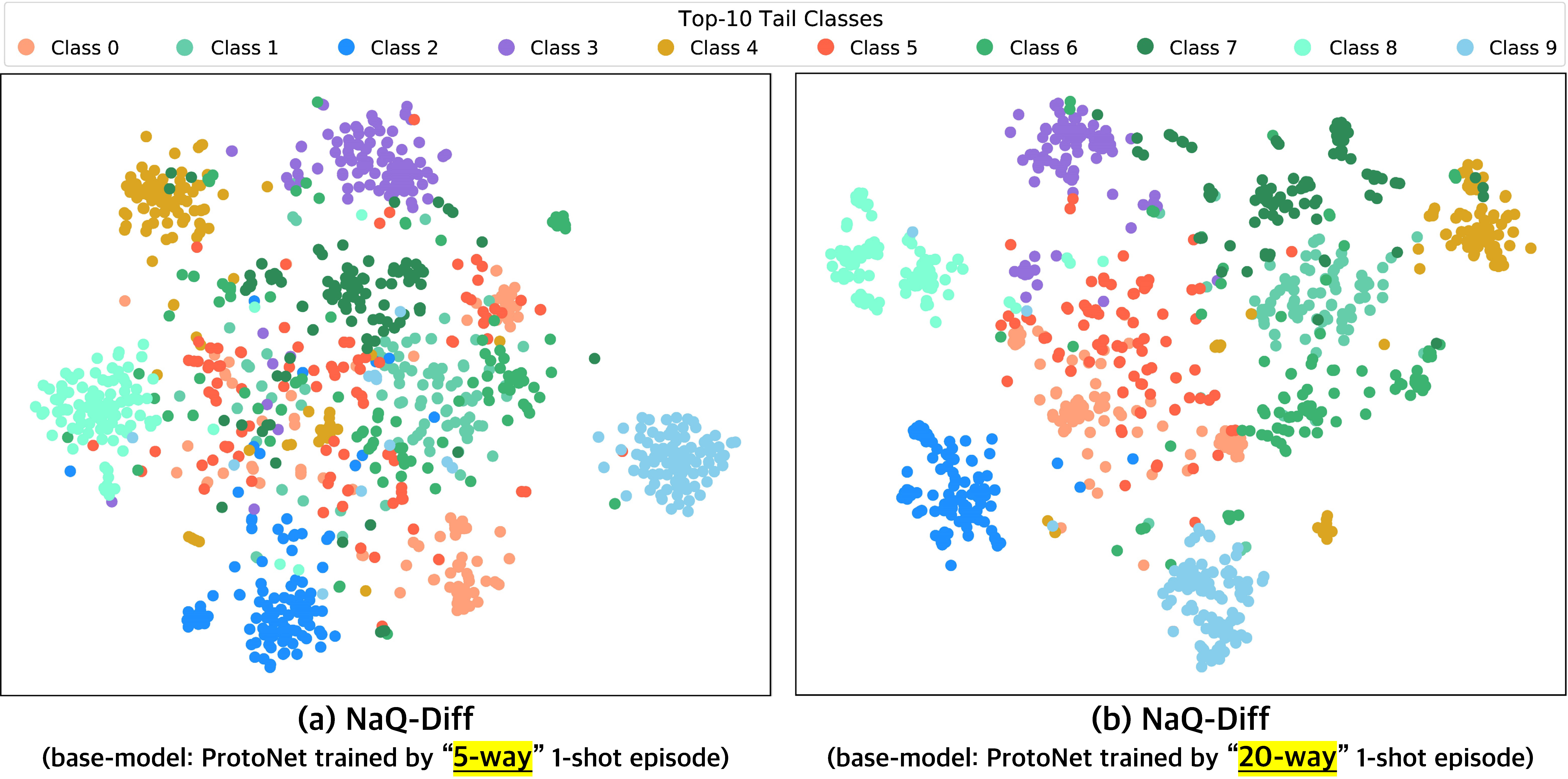}
    \caption{(Left): The t-SNE plot of tail-class embeddings produced by \textsc{NaQ-Diff} trained with \textit{5-way} training episodes. (Right): The t-SNE plot of tail-class embeddings produced by \textsc{NaQ-Diff} trained with \textit{20-way} training episodes (base-model: ProtoNet)}
    \label{class_imbal_tsne:diff_5vs20}
\end{figure}

As we observed in Figure~\ref{class_imbal_tsne:ae}, \textsc{NaQ-Diff} has difficulty in finding class-level similar queries (See Figure~\ref{analysis:classsim} and Table~\ref{imbal:discuss_table}) due to the low average degree ($\sim$2.06) of the Amazon-Electronics dataset, so that produces inferior tail-class embedding quality compared to \textsc{NaQ-Feat} in case of Amazon-Electronics. However, by training with more challenging episodes (i.e., 20-way training episodes), \textit{\textsc{NaQ-Diff} can produce clearly separable tail-class node embeddings even in the Amazon-Electronics dataset.} Therefore, we can conclude that downstream task `format' information provided by episodic learning benefits learning about minority classes.

\begin{figure} [ht!]
    \centering
        \subfigure[Structure noise]{
        \includegraphics[width=0.35\columnwidth]{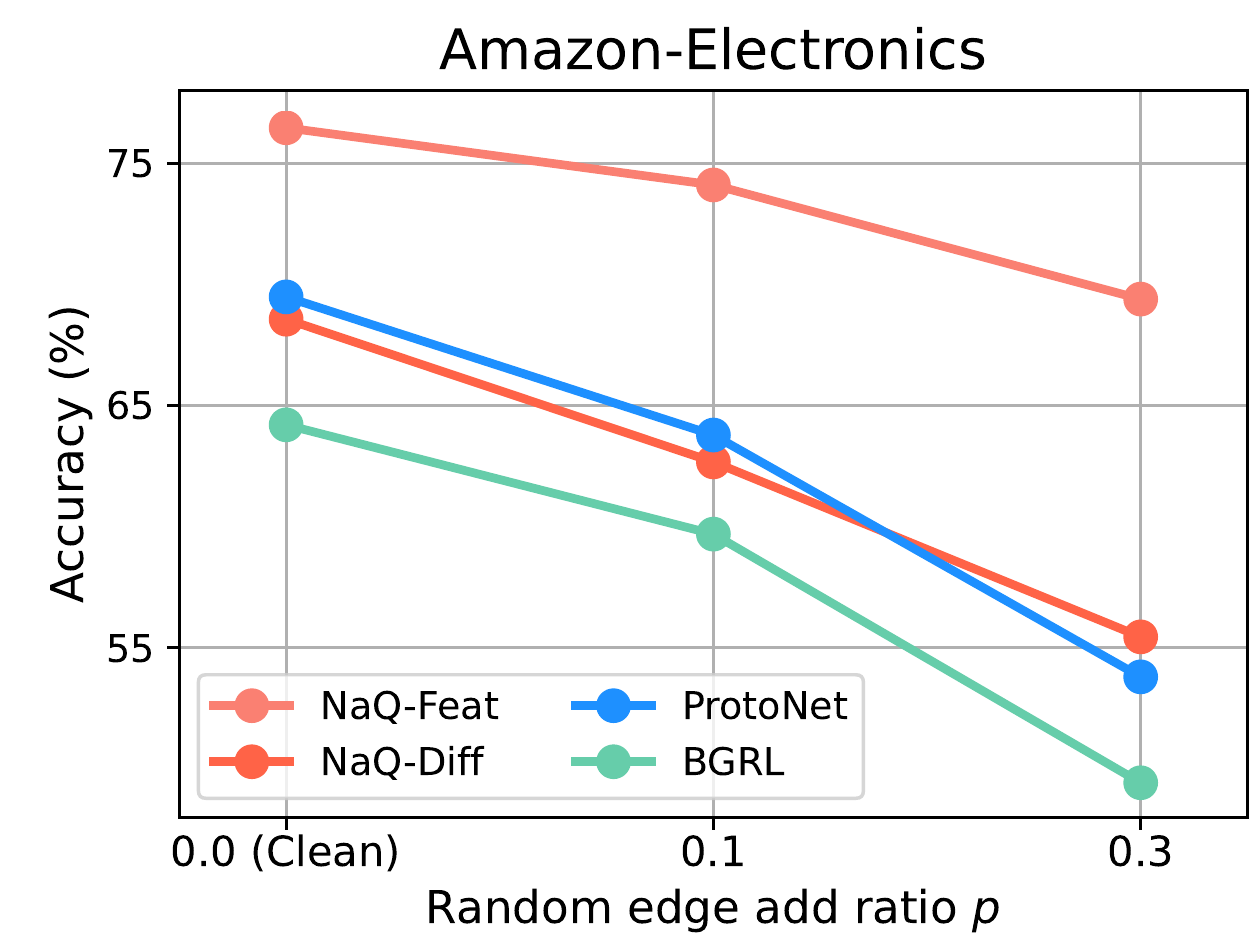}
        \label{bias:struc_noise}
        \vspace{-1mm}
        }
        \hspace{1mm}
        \subfigure[Feature noise]{
        \includegraphics[width=0.35\columnwidth]{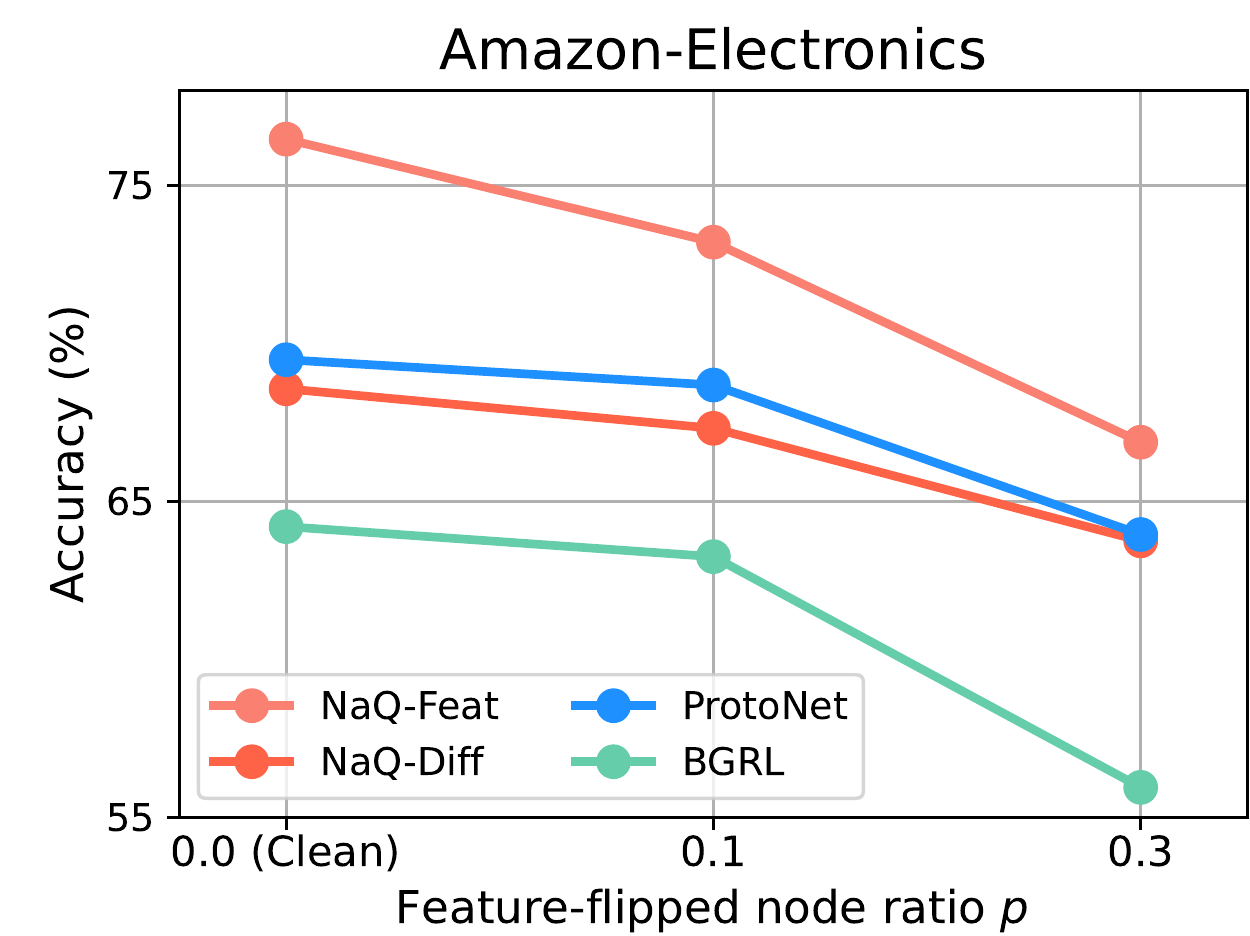}
        \label{bias:feat_noise}
        \vspace{-1mm}
        }
    \vspace{-3mm}
    \caption{(a): Impact of the structure noise, (b): Impact of the feature noise (5-way 1-shot, \textsc{NaQ} base-model: ProtoNet)}
    \label{appendix:bias:fig}
\end{figure}

\subsubsection{Additional results on the Inherent Bias} \label{appendix:bias:result}
\textbf{Structure Noise. } Since structure noise in a graph is also a crucial inherent bias that 
is known to deteriorate the performance of GNNs, we also evaluated the FSNC performance when there are noisy edges in the given graph structure. To perturb graph structure, we considered random edge addition, because adding edges are known to be a more effective attack~\cite{wu2019adversarial}. We add random edges as much as $p\in\{0.1, 0.3\}$ of the number of edges in the original graph. By adjusting the random edge adding ratio $p$, we examined the impact of the structure noise on 5-way 1-shot FSNC performance. Results are presented in Figure~\ref{bias:struc_noise}. We observe that meta-learning methods are more robust than a GCL method, BGRL, which we attribute to the task format information learned by episodic learning framework. Moreover, \textsc{NaQ-Feat} shows significantly better robustness compared with other baselines, as it only utilizes clean raw node feature instead of noisy structure for training episode generation. 

\textbf{Feature Noise. } We also examined the impact of the feature noise on the FSNC performance. After random sampling $p\in\{0.1, 0.3\}$ nodes to be corrupted~\cite{liu2021graph}, we injected feature noise into sampled nodes by randomly flipping 0/1 value on each dimension of the node feature $X_{:.i}$ from Bernoulli distribution with probability $\frac{1}{d}\sum_{i=1}^{d}X_{:.i}$~\cite{zhang2020feature, huang2023robust}. By adjusting the ratio of noisy nodes, we examined the impact of noisy features on 5-way 1-shot FSNC performance. Results are presented in Figure~\ref{bias:feat_noise}. We observe that as more noise is added, BGRL shows a significant performance drop compared to meta-learning methods except for \textsc{NaQ-Feat}, which we attribute again to the task format information learned by episodic learning framework. As expected, as \textsc{NaQ-Feat} relies on the node features for the similarity computation, its performance drops as more feature noise is added.
Thus, developing a more robust algorithm under feature noise will be a promising direction for future work.

\subsection{Ablation Study: Similarity Metric in \textsc{NaQ-Feat}} \label{sim:measure}
As discussed in the Section~\ref{naq:explain}, the choice of the similarity metric is an important factor for \textsc{NaQ-Feat}, since inappropriate choice of the similarity metric can lead to the wrong selection of queries.
To examine the impact of the similarity metric, we use the cosine similarity and the negative Euclidean distance to measure the class-level similarity between each node and top-10 similar nodes found by \textsc{NaQ-Feat} (Table~\ref{table:sim:meas1}), as done in Section~\ref{analysis:empirical}.
Note that Jaccard similarity is excluded when measuring class-level similarity since it cannot be applied to the continuous features.
In addition, we evaluated the 5-way 1-shot FSNC performance on each dataset when using the cosine similarity, Jaccard similarity, and the negative Euclidean distance as the similarity metric (Table~\ref{table:sim:meas2}). Note that all hyperparameter settings of \textsc{NaQ-Feat} other than the similarity metric are identical.

In Table~\ref{table:sim:meas1}, we observe that using the cosine similarity as the similarity metric discovers more class-level similar nodes than using the negative Euclidean distance. As a result, in Table~\ref{table:sim:meas2}, the FSNC accuracy when using the cosine similarity is superior to when using the negative Euclidean distance. Note that this is mainly due to the fact that the datasets used in this experiment have bag-of-words node features, and thus the cosine similarity serves as a better metric.
Therefore, we can confirm that choosing an appropriate  similarity metric is important.

\begin{table}[h!]
\centering
\caption{Impact of the similarity metric on class-level similarity between each node and top-10 similar nodes found via \textsc{NaQ-Feat}.}
\vspace{1mm}
\label{table:sim:meas1}
\renewcommand{\arraystretch}{1.2}
\resizebox{0.62\columnwidth}{!}{%
\begin{tabular}{c|c|c}
\hline
\begin{tabular}[c]{@{}c@{}}Datasets\\ (Feature type: bag-of-words)\end{tabular} & \begin{tabular}[c]{@{}c@{}}Avg. Class-level sim.\\ (Cosine sim.)\end{tabular} & \begin{tabular}[c]{@{}c@{}}Avg. Class-level sim.\\ (Neg. Euclidean dist.)\end{tabular} \\ \hline
Amazon-Clothing    & $\sim$ 81.6\% & $\sim$ 61.0\% \\
Amazon-Electronics & $\sim$ 81.9\% & $\sim$ 64.6\% \\
Cora-Full          & $\sim$ 79.8\% & $\sim$ 40.4\% \\
DBLP               & $\sim$ 73.5\% & $\sim$ 19.1\% \\ \hline
\end{tabular}%
}
\vspace{-3ex}
\end{table}

\begin{table}[h!]
\centering
\caption{Impact of the similarity metric on \textsc{NaQ-Feat} (5-way 1-shot, base-model: ProtoNet)}
\vspace{1mm}
\label{table:sim:meas2}
\renewcommand{\arraystretch}{1.2}
\resizebox{0.71\columnwidth}{!}{%
\begin{tabular}{c|c|c|c}
\hline
\begin{tabular}[c]{@{}c@{}}Datasets\\ (Feature type: bag-of-words)\end{tabular} & \begin{tabular}[c]{@{}c@{}}FSNC Accuracy\\ (Cosine sim.)\end{tabular} &
\begin{tabular}[c]{@{}c@{}}FSNC Accuracy\\ (Jaccard sim.)\end{tabular} &
\begin{tabular}[c]{@{}c@{}}FSNC Accuracy\\ (Neg. Euclidean dist.)\end{tabular} \\ \hline
Amazon-Clothing    & 83.77\% & 83.35\% & 80.83\% \\
Amazon-Electronics & 76.46\% & 76.63\% & 70.68\% \\
Cora-Full          & 64.20\% & 63.53\% & 45.60\% \\
DBLP               & 71.38\% & 72.68\% & 67.53\% \\ \hline
\end{tabular}%
}
\end{table}

When comparing cosine similarity and Jaccard similarity, since they are similar metrics when measuring similarities in bag-of-words data, \textsc{NaQ-Feat} with both similarity metrics shows similar FSNC performance over four datasets as shown in Table~\ref{table:sim:meas2}. Thus, we have the freedom to choose one of those two metrics when using \textsc{NaQ-Feat} on data with bag-of-words features. However, Jaccard similarity cannot be computed with continuous features, as we mentioned above. Hence, it will be more beneficial to consider cosine similarity as a similarity metric due to its generality.

Lastly, note that we did not consider the learnable similarity metric since it requires node-node similarity calculation process per model update for episode generation, which is computationally burdensome. For this reason, we have not considered the learnable metric since we pursued an episode generation method that can be performed \textit{before the training phase.}

\begin{figure} [ht!]
  \centering
  \vspace{1ex}
  \includegraphics[width=0.7\columnwidth]{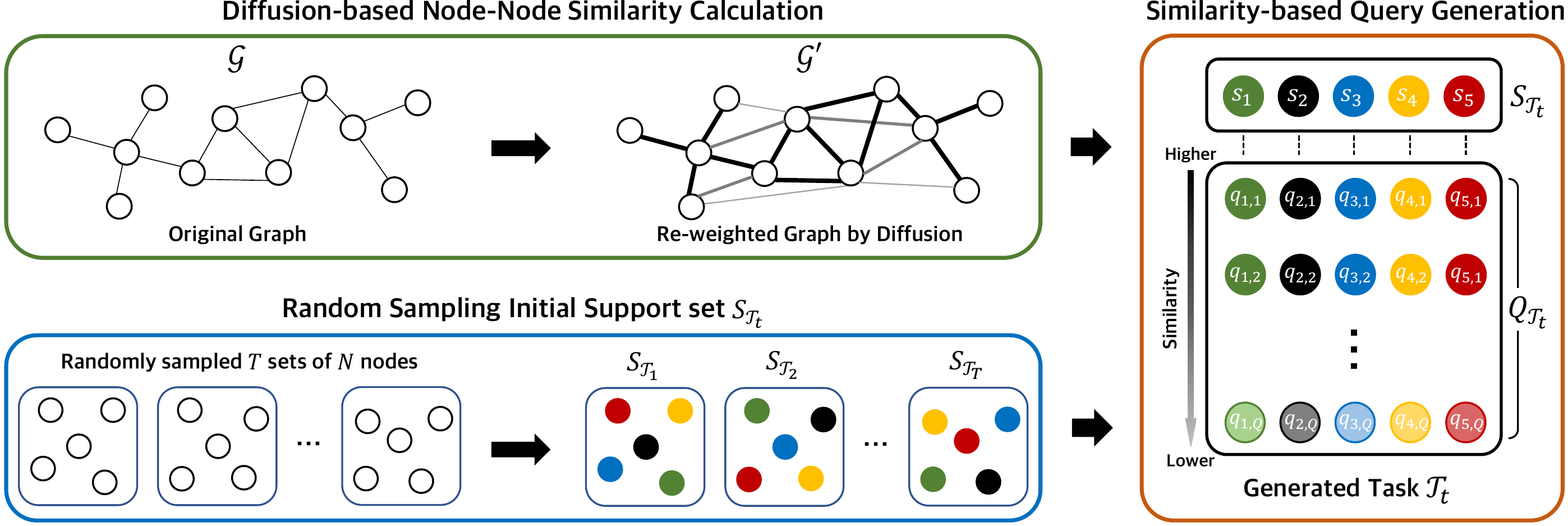}
  \vspace{-3mm}
  \caption{Overview of the \textsc{NaQ-Diff}. The only difference from \textsc{NaQ-Feat} is that \textsc{NaQ-Diff} utilizes graph diffusion instead of raw-feature-based similarity to get node-node similarity.}
  \label{naq_diff:overview}
  \vspace{-2ex}
\end{figure}

\subsection{Details on \textsc{NaQ-Diff}} \label{naq_diff}
For \textsc{NaQ-Diff}, we used Personalized PageRank (PPR)~\cite{ppr}-based diffusion to obtain diffusion matrix $\mathbf{S}$, where $\theta^{PPR}_k=\alpha(1-\alpha)^k$, with teleport probability $\alpha \in (0,1)$, as the weighting coefficient $\theta_k$. In our experiments, $\alpha=0.1$ is used to calculate PPR-based diffusion. Also, we used $\tilde{T}_{sym}=(w_{loop}\cdot\mathbf{I}_N+D)^{-1/2}(w_{loop}\cdot\mathbf{I}_N+A)(w_{loop}\cdot\mathbf{I}_N+D)^{-1/2}$, with the self-loop weight $w_{loop}=1$, as transition matrix, where $A\in\mathbb{R}^{|\mathcal{V}|\times|\mathcal{V}|}$ is an adjacency matrix of the graph $\mathcal{G}$ and $D$ is a diagonal matrix whose entries $D_{ii}=\sum_{j}A_{ij}$ are each node's degree.

Last but not least, although there can be other approaches for capturing the graph structural information (e.g., using the adjacency matrix, or using a k-NN graph computed based on node embeddings learned by a GNN encoder \textit{during the training phase})~\cite{rgrl}, we choose the graph diffusion as it captures more global information than the adjacency matrix, and computationally more efficient than the k-NN approach.

\subsection{Details on Evaluation Datasets} \label{detail:dataset}
The following is the details on evaluation datasets used in this work. \vspace{-1ex}
\begin{itemize}
    \item \textbf{Amazon-Clothing}~\cite{amazonnetwork} is a product-product network, whose nodes are products from the category ``Clothing, Shoes and Jewelry'' in Amazon. Node features are constructed from the product descriptions, and edges were created based on "also-viewed" relationships between products. The node class is a low-level product category.
    \item \textbf{Amazon-Electronics}~\cite{amazonnetwork} is a network of products, whose nodes are products from the category ``Electronics'' in Amazon. Node features are constructed from the product descriptions, and edges represent the ``bought-together'' relationship between products. The node class is a low-level product category.
    \item \textbf{Cora-Full}~\cite{corafull} is a citation network, whose nodes are papers. The node features are constructed from a bag-of-words representation of each node's title and abstract, and edges represent the citation relationship between papers. The node class is the paper topic.
    \item \textbf{DBLP}~\cite{dblp} is a citation network, whose nodes are papers. Node features are constructed from their abstracts, and edges represent the citation relationship between papers. The node class is the venue where the paper is published.
    \item \textbf{ogbn-arxiv}~\cite{hu2020ogb} is a citation network, whose nodes are CS arXiv papers. Node features are constructed by averaging the embeddings of words in the title and abstract, where the word embeddings are obtained from the skip-gram model~\cite{skipgram} over the MAG~\cite{mag} corpus. Edges are citation relationships between papers, and the node class is 40 subject areas of arXiv CS papers.
\end{itemize}

The detailed statistics of the datasets can be found in Table~\ref{datastat}. ``Hom. ratio'' denotes the homophily ratio of each dataset, and ``Class split'' denotes the number of distinct classes used to generate episodes in training (\textit{only for supervised settings}), validation, and testing phase, respectively.
For ogbn-arxiv, due to the GPU memory issue, graph diffusion calculation is done as a truncated sum. Moreover, as node features in ogbn-arxiv are word embeddings, we used the negative Euclidean distance as the similarity metric used for sampling query nodes in \textsc{NaQ}.

\begin{table}[h!]
\caption{Dataset statistics.}
\centering
\vspace{1mm}
\label{datastat}
\renewcommand{\arraystretch}{1.2}
\resizebox{0.7\columnwidth}{!}{%
\begin{tabular}{ccccccc}
\hline
Dataset            & \# Nodes & \# Edges & \# Features & \# Labels & Class split & Hom. ratio \\ \hline
Amazon-Clothing    & 24,919  & 91,680  & 9,034      & 77       & 40/17/20    & 0.62 \\
Amazon-Electronics & 42,318  & 43,556  & 8,669      & 167      & 90/37/40    & 0.38 \\
Cora-Full          & 19,793  & 65,311  & 8,710      & 70       & 25/20/25    & 0.59 \\
DBLP               & 40,672  & 288,270 & 7,202      & 137      & 80/27/30    & 0.29 \\ 
ogbn-arxiv         & 169,343 & 1,166,243 & 128      & 40       & 15/10/15    & 0.43 \\ \hline
\end{tabular}%
}
\vspace{-1ex}
\end{table}

\subsection{Details on Compared Baselines \& Experimental Settings} \label{detail:baseline}
Details on compared baselines are presented as follows.
\begin{itemize}
    \item \textbf{MAML}~\cite{maml} aims to find good initialization for downstream tasks. It optimizes parameters via two-phase optimization. The inner-loop update finds task-specific parameters based on the support set of each task, and the outer-loop update finds a good parameter initialization point based on the query set.
    \item \textbf{ProtoNet}~\cite{protonet} trains a model by building $N$ class prototypes by averaging support samples of each class, and making each query sample and corresponding prototype closer.
    \item \textbf{G-Meta}~\cite{gmeta} obtains node embeddings based on the subgraph of each node in episodes, which allows scalable and inductive graph meta-learning.
    \item \textbf{TENT}~\cite{tent} performs graph meta-learning to reduce the task variance among training episodes via node-level, class-level, and task-level adaptations.
    \item \textbf{GLITTER}~\cite{glitter} aims to learn task-specific structures consisting of support set nodes and their relevant nodes, which have high node influence on them for each meta-training/test task since the given original graph structure is redundant when learning node embeddings in each meta-task.
    \item \textbf{COSMIC}~\cite{cosmic} adopts contrastive learning scheme on graph meta-learning to enhance the intra-class generalizability and similarity-sensitive mix-up which generates hard (unseen) node classes for the inter-class generalizability.
    \item \textbf{BGRL}~\cite{bgrl} applies BYOL~\cite{byol} on graphs, so it trains the model by maximizing the agreement between an online embedding and a target embedding of each node, where each embedding is obtained from two differently augmented views.
    \item \textbf{SUGRL}~\cite{sugrl} simplifies architectures for effective and efficient contrastive learning on graphs, and trains the model by concurrently increasing inter-class variation and reducing intra-class variation.
    \item \textbf{AFGRL}~\cite{afgrl} applies BYOL architecture without graph augmentations. Instead of augmentations, AFGRL generates another view by mining positive nodes in the graph in terms of both local and global perspectives.
    \item \textbf{VNT}~\cite{vnt} utilizes pretrained transformer-based encoder (Graph-Bert~\cite{graph-bert}) as a backbone, and adapts to the downstream FSNC task by tuning injected `virtual' nodes and classifier with given a few labeled samples in the downstream task, then makes prediction on queries with such fine-tuned virtual nodes and classifier.
\end{itemize}

For meta-learning baselines except for GLITTER and COSMIC, we used a 2-layer GCN~\cite{gcn} as the GNN encoder with the hidden dimension chosen from \{64, 256\}, and this makes MAML to be essentially equivalent to Meta-GNN~\cite{metagnn}. Such choice of high hidden dimension size is based on~\cite{chen2019closer}, which demonstrated that a larger encoder capacity leads to a higher performance of meta-learning model. For each baseline, we tune hyperparameters for each episode generation method.
In the case of GLITTER\footnote{https://github.com/SongW-SW/GLITTER} and COSMIC\footnote{https://github.com/SongW-SW/COSMIC}, we adopted the settings regarding the GNN encoder (e.g., number of layers and GNN model type) and hyperparameter settings reported in their official source code.
For GCL baselines, we also used a 2-layer GCN encoder with the hidden dimension of size 256. For VNT, following the original paper, Graph-Bert~\cite{graph-bert} is used as the backbone transformer model. As the official code of VNT is not available, we tried our best to reproduce VNT with the settings presented in the paper of VNT and Graph-Bert.
For all baselines, Adam~\cite{adam} optimizer is used. The tuned parameters and their ranges are summarized in Table~\ref{hyparam}.
{Note that training TENT with \textsc{NaQ} was non-trivial, as it utilizes the entire labeled data $(X_{C_b},Y_{C_b})$ to compute cross-entropy loss along with episode-specific losses computed with training episodes per each update. Therefore, when we train TENT with \textsc{NaQ}, the cross-entropy loss was calculated over a single episode. For this reason, the superior performance of \textsc{NaQ} with TENT is especially noteworthy (See Figure~\ref{naq_model_agnostic} and~\ref{naq_model_agnostic_higher_way}) as it outperforms vanilla supervised TENT even with much less data involved in each parameter update during the training phase.}

\begin{table}[ht]
\centering
\vspace{-1ex}
\caption{Tuned hyperparameters and their range by baselines}
\vspace{1mm}
\label{hyparam}
\renewcommand{\arraystretch}{1.2}
\resizebox{0.6\columnwidth}{!}{%
\begin{tabular}{c|clllll}
\hline
Baselines                                                                & \multicolumn{6}{c}{Hyperparameters and Range}                                                                                                                                                                      \\ \hline
\begin{tabular}[c]{@{}c@{}}MAML-like\\ (MAML, G-Meta)\end{tabular}       & \multicolumn{6}{c}{\begin{tabular}[c]{@{}c@{}}Inner step learning rate $\in\{0.01, 0.05, 0.1, 0.3, 0.5\}$,\\ \# of inner updates $\in\{1, 2, 5, 10, 20\}$, Meta-learning rate $\in\{0.001, 0.003\}$\end{tabular}} \\ \hline
\begin{tabular}[c]{@{}c@{}}ProtoNet-like\\ (ProtoNet, TENT)\end{tabular} & \multicolumn{6}{c}{Learning rate $\in\{5\cdot 10^{-5}, 10^{-4}, \  3\cdot 10^{-4}, 5\cdot 10^{-4}, 10^{-3}, 3\cdot 10^{-3}, 5\cdot 10^{-3} \}$}                                                                      \\ \hline
Self-Supervised (TLP)                                                    & \multicolumn{6}{c}{Learning rate $\in\{10^{-6}, 10^{-5}, 5\cdot 10^{-5}, 10^{-4}, 5\cdot 10^{-4}, 10^{-3}\}$}                                                                                                                                      \\ \hline
\end{tabular}%
}
\end{table}

\begin{figure} [ht!]
    \centering
    \includegraphics[width=0.75\columnwidth]{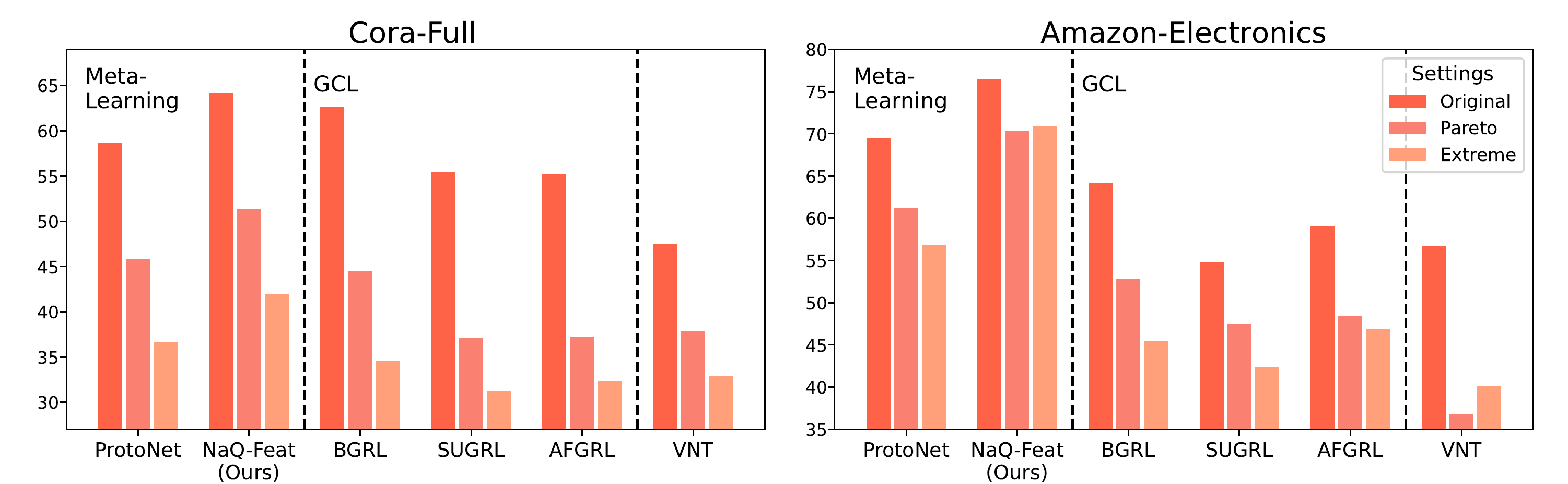}
    \vspace{-4mm}
    \caption{Impact of the class imbalance (5-way 1-shot, \textsc{NaQ-Feat} base-model: ProtoNet)}
    \label{appendix:class_imbal}
\end{figure}

\textbf{Discussion on VNT. }
Although we tried our best to reproduce VNT, we failed to achieve their reported performance especially on Cora-Full, an evaluation dataset shared by VNT and our paper. This might be due to the random seed, dataset split, or the transformer architecture used in the experiment. However, we conjecture it will also suffer from the inherent bias in data such as class imbalance similar to GCL methods, as graph transformer-based model also learn generic embedding by pretraining on a given graph. 
As an evidence, in Figure~\ref{appendix:class_imbal}, we show the results on Cora-Full and Amazon-Electronics under the same setting used to report results in Figure~\ref{intro:class_imbal}. We observe that the performance of VNT deteriorates under class imbalance like GCL methods.

\subsection{Model Training with Episodes from NaQ: ProtoNet Example} \label{appx:training:proto}
In this section, we explain how to train ProtoNet~\cite{protonet}, which is one of the most widely used meta-learning models, with episodes generated by \textsc{NaQ}, as a detailed example of Algorithm~\ref{alg:naq:train}. Let $f_\theta$ be a GNN encoder, $\mathcal{T}_t$ be a generated episode, and $S_{\mathcal{T}_t}=\{(x^{spt}_{t,i}, y^{spt}_{t,i})\}_{i=1}^{N\times K}$ be a randomly sampled support set, then a corresponding query set is generated as $Q_{\mathcal{T}_t}=\{(x^{qry}_{t,i}, y^{qry}_{t,i})\}_{i=1}^{N\times Q} = \text{\textsc{NaQ}}(S_{\mathcal{T}_t})$. 

More precisely, we first obtain a prototype $\mathbf{c}_j$ for each class $j\in\{1,\cdots, N\}$ based on the support set $S_{\mathcal{T}}$ as follows\footnote{To remove clutter, we drop the task subscript $t$ from all notations from now on.}: 
\begin{equation} \label{prototype}
    \mathbf{c}_j = \frac{1}{K} \sum\nolimits_{i=1}^{K} f_{\theta}(x_{i}^{spt}) \cdot \mathbb{I} \{y_i^{spt}=j \}  
\end{equation}
where $\mathbb{I} \{y_i^{spt}=j \}$ is an indicator function that is equal to 1 only if the label $y_i$ of $x_i$ is $j$, otherwise 0. 
Then, the probability of each query $(x^{qry},y^{qry})\in Q_\mathcal{T}$ belonging to class $j$ is computed as follows:
\begin{equation}\label{proto-query}
    P(y^{qry}=j;x^{qry}) = \frac{\mathrm{exp}(-d(f_{\theta}(x^{qry}), \ \mathbf{c}_j))}{\sum_{j'}\mathrm{exp}(-d(f_{\theta}(x^{qry}), \ \mathbf{c}_{j'}))} 
\end{equation}
where $d(\cdot,\cdot)$ is a distance function. We use Euclidean distance in this work.

Then, the parameter is updated as: $\theta \leftarrow \theta - \eta \nabla_{\theta} \mathcal{L}(\theta;qry)$, where $\eta$ is the learning rate and $\mathcal{L}(\theta;qry)$ is a loss given as: 
\begin{equation}
    \mathcal{L}(\theta;qry) = -\frac{1}{N\times Q}\sum\nolimits_{(x^{qry},y^{qry})\in Q_\mathcal{T}}\mathrm{log}(P(y^{qry}=j;x^{qry})). 
\end{equation}

\subsection{Discussion on the Time Complexity of \textsc{NaQ}}
In this section, we provide the time analysis of \textsc{NaQ} for generating training episodes. 
We measured the time spent for the similarity calculation in  each dataset, and the time taken to generate all training episodes (i.e., 16,000 in total). The results can be found in Table~\ref{time:simcal} and~\ref{time:epigen}. Even though the datasets we used are not small, NaQ does not require significant time costs. Moreover, when we use \textsc{NaQ}, the time cost required for similarity calculation and episode generation is at least three times faster than for ordinary supervised methods’ training episode generation.

\begin{table}[h!]
\centering
\caption{Averaged elapsed time over 5 runs in seconds for node-node similarity calculation.}
\vspace{1mm}
\label{time:simcal}
\resizebox{0.4\columnwidth}{!}{%
\begin{tabular}{c|cc}
\hline
Dataset            & \textsc{NaQ-Feat} & \textsc{NaQ-Diff} \\ \hline
Amazon-Clothing    & 1.7769            & 2.7194            \\
Amazon-Electronics & 0.6538            & 11.0443           \\
Cora-Full          & 0.0014            & 1.3207            \\
DBLP               & 0.0194            & 9.8653            \\ \hline
\end{tabular}%
}
\end{table}

\begin{table}[h!]
\centering
\vspace{-3ex}
\caption{Averaged elapsed time over 5 runs in seconds for generating 16,000 training episodes.}
\vspace{1mm}
\label{time:epigen}
\resizebox{0.5\columnwidth}{!}{%
\begin{tabular}{c|cc|c}
\hline
Dataset            & \textsc{NaQ-Feat} & \textsc{NaQ-Diff} & Supervised \\ \hline
Amazon-Clothing    & 5.2117            & 5.0242            & 64.0850    \\
Amazon-Electronics & 6.1301            & 5.9622            & 64.4830    \\
Cora-Full          & 4.9786            & 4.7484            & 57.2931    \\
DBLP               & 5.9689            & 6.5407            & 65.8119    \\ \hline
\end{tabular}%
}
\end{table}

However, there are cases where it is challenging to perform similarity calculations at once due to GPU memory problems, if the size of the graph is too large.
In such cases, we can calculate the node-node similarity by performing node-wise calculation (\textsc{NaQ-Feat}) or calculating graph diffusion as a truncated sum (\textsc{NaQ-Diff}), where this process is required only once for each dataset.
Then, a list of Top-$k$ ($k<<$ \# of nodes) similar nodes for each node can be stored and used by loading them during the episode generation process. For example, in the case of the ogbn-arxiv dataset, which contains about 160,000 nodes, we can calculate the Top-$k$ similar nodes list with a capacity of $\sim$129.20 MiB in a short time of about 150 seconds for \textsc{NaQ-Feat} and 740 seconds for \textsc{NaQ-Diff}, for $k=100$. By using this Top-$k$ similar nodes list, only 2.2713 for \textsc{NaQ-Feat} and 2.2266 for \textsc{NaQ-Diff} seconds are spent on average (5 times) for generating total 16,000 training episodes, which is faster than the supervised models’ average of 55.9989 seconds in the ogbn-arxiv dataset.

Moreover, in the case of ogbn-products having 2,449,029 nodes, which is a very large-scale dataset, we can calculate such Top-$100$ similar nodes list in a short time of about 705 seconds by using batched node-node similarity calculation. Thus, our \textsc{NaQ-Feat} can be scalable to very large graphs having million scale nodes. The following results presented in Table~\ref{table:ogbn:products} demonstrate the effectiveness of our \textsc{NaQ-Feat} in a very large-scale dataset.

\begin{table}[h!]
    \centering
    \caption{Overall averaged FSNC accuracy (\%) with 95\% confidence intervals on very large-scale dataset (ogbn-products: having 2,449,029 nodes, 61,859,140 edges, 47 classes (class split: 15/15/17), \# of features: 100 (obtained by PCA on bag-of-words features), \textsc{NaQ-Feat} base-model: ProtoNet)}
    \vspace{1mm}
    \renewcommand{\arraystretch}{1.2}
\resizebox{0.4\columnwidth}{!}{%
\begin{tabular}{c|cc}
\hline
Dataset         & \multicolumn{2}{c}{\textbf{ogbn-products}}                                                                \\ \hline
Baselines       & 5-way 1-shot             & 10-way 1-shot \\ \hline
ProtoNet (Sup.) & 43.50±\tiny{1.20} & 34.19±\tiny{0.69}  \\
COSMIC (Sup.)   & OOM & OOM  \\ \hline
TLP-BGRL        & OOM & OOM  \\
TLP-SUGRL       & 27.81±\tiny{0.78} & 18.72±\tiny{0.52}  \\ \hline
\textbf{\textsc{NaQ-Feat} (Ours)} & \textbf{53.82±\tiny{1.26}} & \textbf{43.84±\tiny{0.77}}  \\ \hline
\end{tabular}%
}    
\label{table:ogbn:products}
\end{table}

It is worth noting that \textit{we only need one similarity calculation per dataset}, which makes \textsc{NaQ} practical in reality.

\subsection{Regarding Overlapping Queries in \textsc{NaQ}} \label{qry:overlap}
In this section, we discuss the query overlapping problem of \textsc{NaQ}, where sampled query sets corresponding to each distinct support set have an intersection, which might hurt the FSNC performance of \textsc{NaQ}.
Although we tried to prevent this problem by generating only a 1-shot support set as we mentioned in ‘Support set generation’ process in Section~\ref{naq:explain} (In other words, as each class contains only a 1-shot support node, the number of overlapping queries among classes can be minimized.), such query overlapping problem can happen and might be problematic for \textsc{NaQ}.
To assess the severity of this problem, we measured the average query overlap ratio within training episodes generated by NaQ for each dataset. As shown in Table~\ref{table:qry:overlap} below, query overlap is generally very rare case.

\begin{table}[h!]
\centering
\vspace{-2ex}
\caption[Averaged query overlap ratio within training episodes generated by \textsc{NaQ}]{Averaged query overlap ratio within 16,000 training episodes generated by \textsc{NaQ}}
\vspace{1mm}
\label{table:qry:overlap}
\renewcommand{\arraystretch}{1.15}
\resizebox{0.95\columnwidth}{!}{%
\begin{tabular}{c|cc|cc|cc|cc}
\hline
Datasets &
  \multicolumn{2}{c|}{\textbf{Amazon-Clothing}} &
  \multicolumn{2}{c|}{\textbf{Amazon-Electronics}} &
  \multicolumn{2}{c|}{\textbf{Cora-Full}} &
  \multicolumn{2}{c}{\textbf{DBLP}} \\ \hline
$N$-way &
  \textsc{NaQ-Feat} &
  \textsc{NaQ-Diff} &
  \textsc{NaQ-Feat} &
  \textsc{NaQ-Diff} &
  \textsc{NaQ-Feat} &
  \textsc{NaQ-Diff} &
  \textsc{NaQ-Feat} &
  \textsc{NaQ-Diff} \\ \hline
5    & 0.1573\% & 0.9978\% & 0.0871\% & \underline{11.1715\%} & 0.2206\% & 0.4743\% & 0.1826\% & 0.0605\% \\
10    & 0.3855\% & 2.0769\% & 0.2118\% & \underline{16.9618\%} & 0.5101\% & 1.0138\% & 0.4108\% & 0.1389\% \\
20    & 0.7834\% & 4.0358\% & 0.4457\% & \underline{21.4706\%} & 1.0221\% & 2.0151\% & 0.8559\% & 0.3054\% \\ \hline
\end{tabular}%
}
\end{table}

However, in the Amazon-Electronics dataset, which has a very low average degree ($\sim$2.06), we observe non-negligible overlap ratio in the case of \textsc{NaQ-Diff}, which uses graph Diffusion to find queries. To address this issue, we intentionally dropped overlapping queries in training episodes. Table~\ref{table:drop:diff} and~\ref{table:drop:feat} below show results of the effect of dropping overlapping queries. `Overlap drop ver.' means that we dropped overlapping queries after the episode generation process of \textsc{NaQ}.

\begin{table}[ht]
\centering
\caption{Impact of dropping overlapping queries on FSNC performance (\%) of \textsc{NaQ-Diff} in Amazon-Electronics (base-model: ProtoNet)}
\vspace{1mm}
\label{table:drop:diff}
\renewcommand{\arraystretch}{1.1}
\resizebox{0.43\linewidth}{!}{%
\begin{tabular}{c|cc}
\hline
\multicolumn{3}{c}{\textbf{Amazon-Electronics}} \\ \hline
Setting &
  \begin{tabular}[c]{@{}c@{}}\textsc{NaQ-Diff}\\ \small{(Original ver.)}\end{tabular} &
  \begin{tabular}[c]{@{}c@{}}\textsc{NaQ-Diff}\\ \small{(Overlap drop ver.)}\end{tabular} \\ \hline
5-way 1-shot  & 68.56±\tiny{1.18} & \textbf{69.77±\tiny{1.17}} \\
10-way 1-shot & 59.46±\tiny{0.86} & \textbf{61.98±\tiny{0.86}} \\
20-way 1-shot & 49.24±\tiny{0.59} & \textbf{52.15±\tiny{0.60}} \\ \hline
\end{tabular}%
}
\vspace{-4ex}
\end{table}

\begin{table}[ht]
\centering
\caption{Impact of dropping overlapping queries on FSNC performance (\%) of \textsc{NaQ-Feat} in Cora-Full (base-model: ProtoNet)}
\vspace{1mm}
\label{table:drop:feat}
\renewcommand{\arraystretch}{1.1}
\resizebox{0.43\linewidth}{!}{%
\begin{tabular}{c|cc}
\hline
\multicolumn{3}{c}{\textbf{Cora-Full}} \\ \hline
Setting &
  \begin{tabular}[c]{@{}c@{}}\textsc{NaQ-Feat}\\ \small{(Original ver.)}\end{tabular} &
  \begin{tabular}[c]{@{}c@{}}\textsc{NaQ-Feat}\\ \small{(Overlap drop ver.)}\end{tabular} \\ \hline
5-way 1-shot  & 64.20±\tiny{1.11} & 63.37±\tiny{1.08} \\
10-way 1-shot & 51.78±\tiny{0.75} & 52.32±\tiny{0.75} \\
20-way 1-shot & 40.11±\tiny{0.45} & 40.27±\tiny{0.48} \\ \hline
\end{tabular}%
}
\end{table}

From above results, we can conclude that removing query overlaps is a promising solution when query overlap is not negligible like the case of \textsc{NaQ-Diff} in Amazon-Electronics (see Table~\ref{table:drop:diff}). However, when query overlap is negligible, dropping overlapping queries does not bring remarkable improvements (see Table~\ref{table:drop:feat}).

In summary, the results in Table~\ref{table:qry:overlap} and Table~\ref{table:drop:feat} demonstrate that the query overlapping problem of \textsc{NaQ} is generally negligible in real-world datasets, and the results in Table~\ref{table:drop:diff} imply that dropping overlapping queries can be a promising solution for some of the exceptional cases like \textsc{NaQ-Diff} in Amazon-Electronics dataset.

\subsection{g-UMTRA: Augmentation-based Query Generation Method}
In this section, we introduce our investigation method named g-UMTRA, utilizes graph augmentation to generate queries.
In computer vision, UMTRA~\cite{umtra} tried to apply MAML in an unsupervised manner by generating episodes with image augmentations. With randomly sampled $N$ support set nodes, UMTRA makes a corresponding query set through augmentation on the support set.
Inspired by UMTRA~\cite{umtra}, we devised an augmentation-based query generation method called g-UMTRA. as an investigation method. g-UMTRA generates query set by applying graph augmentation on the support set. The method overview an be found in Figure~\ref{gumtra:overview}.

\begin{figure} [ht!]
  \centering
  \includegraphics[width=0.5\columnwidth]{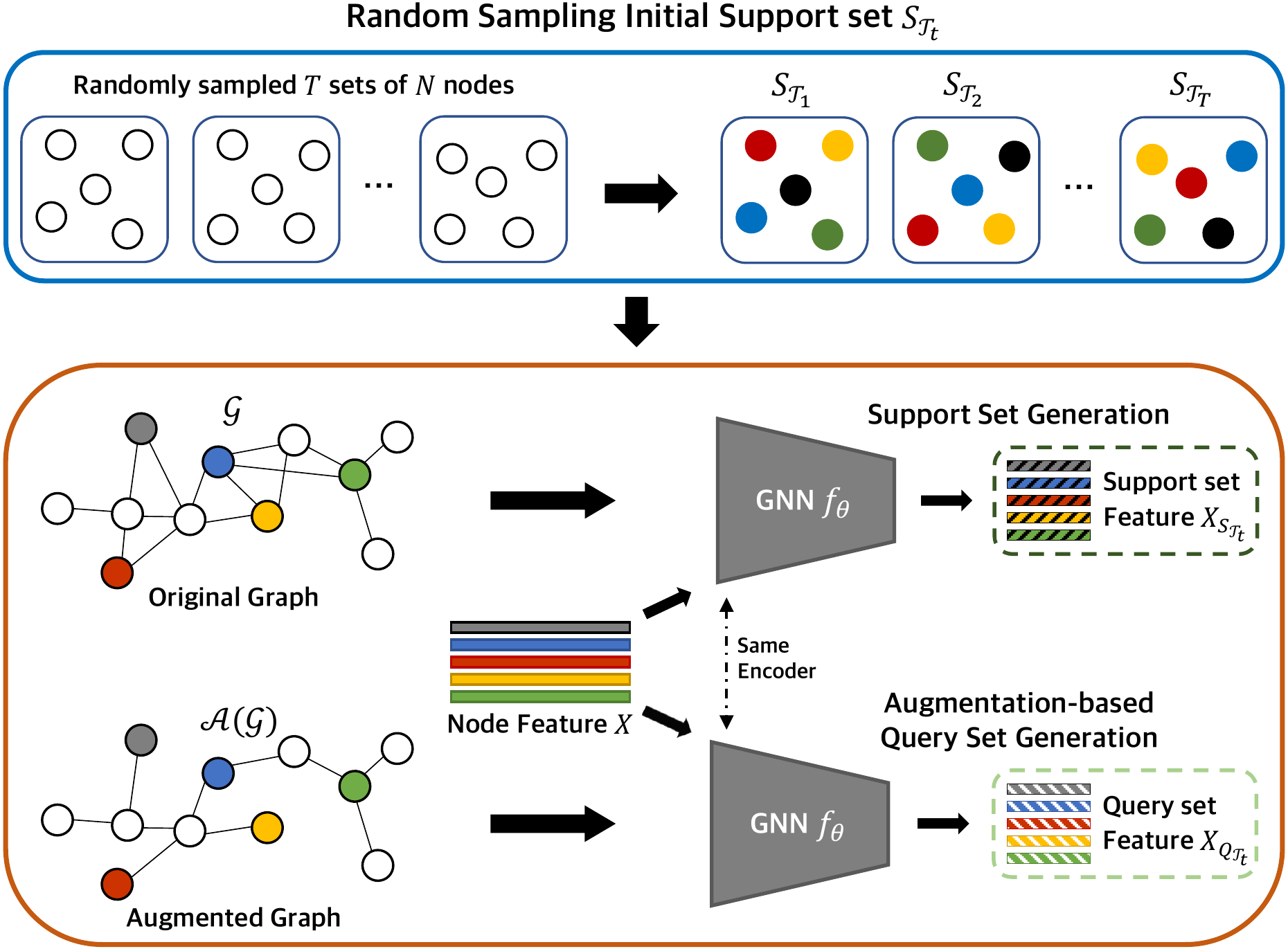}
  \vspace{-3mm}
  \caption{Overview of g-UMTRA. First, we randomly sample $T$ sets of $N$ nodes from the entire graph and assign them distinct labels. Then, we generate support set features by using a GNN encoder with original structures and query set features with augmented structures.}
  \label{gumtra:overview}
\end{figure}

Specifically, we first randomly sample $T$ sets of $N$ nodes from entire graph to generate $\{\mathcal{T}_{t}\}_{t=1}^{T}$. Then for each task $\mathcal{T}_t$, we generate a $N$-way support set $S_{\mathcal{T}_t}=\{ (x_{t,i}, y_{t,i}) \ | \ x_{t,i}\in \mathcal{V} \}_{i=1}^{N\times 1}$ with distinct pseudo-labels $y_{t,i}$ for each $x_{t,i}$, and their corresponding query set $Q_{\mathcal{T}_t}$ in the embedding space by applying graph augmentation. 

By notating GNN encoder $f_\theta$ as $f_{\theta}(\mathcal{V};\mathcal{G})$, we can formally describe the query generation process of g-UMTRA as follows:
\begin{align} \label{gumtra:process}
    X_{S_{\mathcal{T}_t}} & = \{ \big(f_\theta(x_{t,i};\mathcal{G}), y_{t,i}\big) \ | \ (x_{t,i}, y_{t,i})\in S_{\mathcal{T}_t} \}, \nonumber \\
    X_{Q_{\mathcal{T}_t}} & = \{ \big(f_\theta (x_{t,i};\mathcal{A}(\mathcal{G})), y_{t,i}\big) \ | \ (x_{t,i}, y_{t,i})\in S_{\mathcal{T}_t} \},
\end{align}
where $f_\theta(x_{t,i};\mathcal{G})$ is an embedding of node $x_{t,i}$ with the given graph $\mathcal{G}$ and a GNN encoder $f_\theta$, and $\mathcal{A}(\cdot)$ is a graph augmentation function. For $\mathcal{A}(\cdot)$, we can consider various strategies such as node feature masking (DropFeature) or DropEdge~\cite{dropedge}.

Note that g-UMTRA is distinguished from UMTRA in the following two aspects: (1) g-UMTRA can be applied to any existing graph meta-learning methods as it only focuses on episode generation, while UMTRA is mainly applied on MAML. (2) As described in Equation \ref{gumtra:process}, g-UMTRA generates episodes as pair of sets $(X_{S_{\mathcal{T}_j}}, X_{Q_{\mathcal{T}_j}})$ that consist of embeddings. Hence, its query generation process should take place in the training process, since augmentation and embedding calculation of GNNs depend on the graph structure. However, in UMTRA, image augmentation and ordinary convolutional neural networks are applied in instance-level, implying that the episode generation process of UMTRA can be done before training.

\begin{figure} [ht!]
  \centering
  \includegraphics[width=0.6\columnwidth]{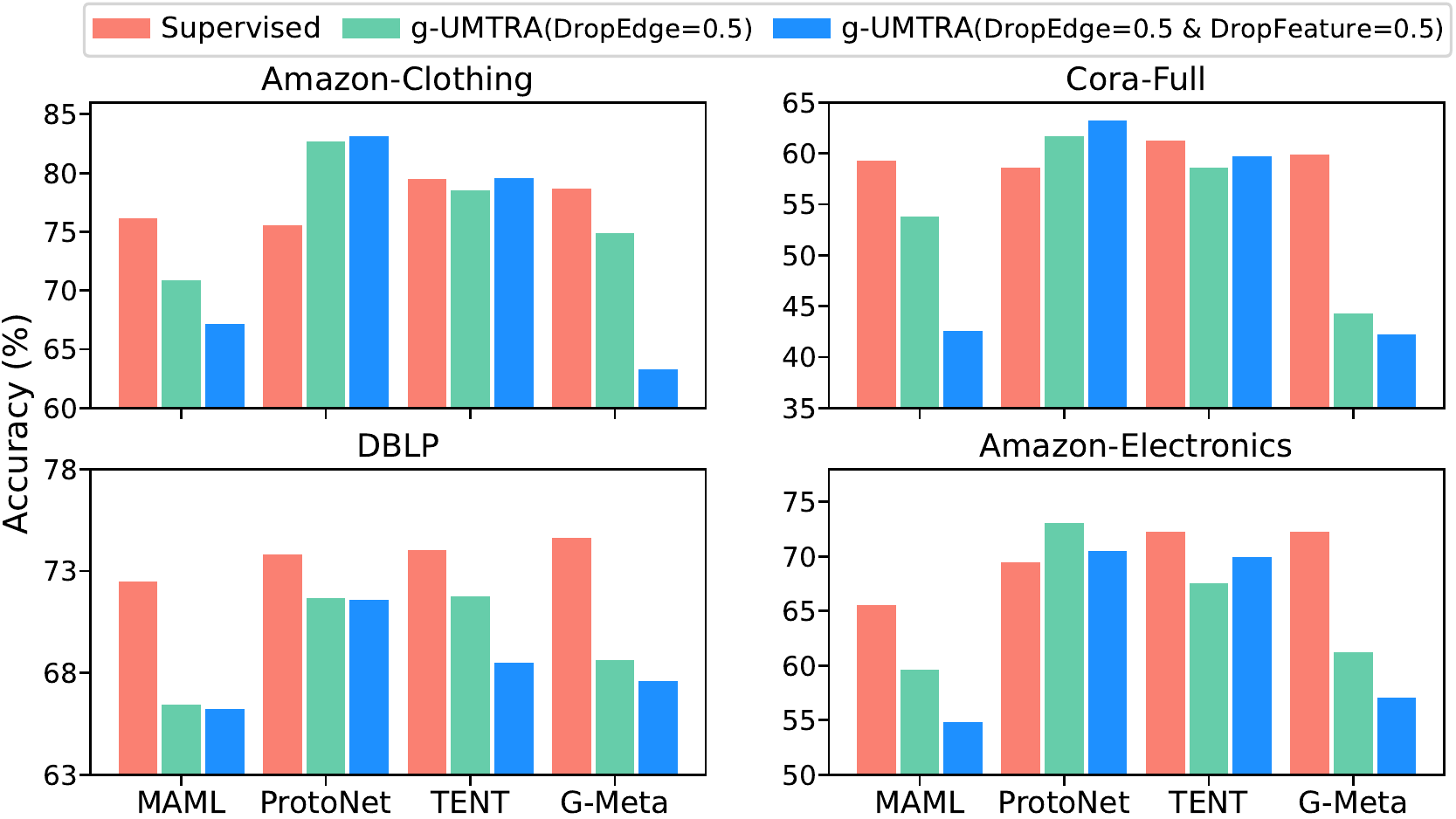}
  \vspace{-1mm}
  \caption{Performance comparison between supervised, g-UMTRA with DropEdge, and g-UMTRA with DropEdge and DropFeature on existing graph meta-learning models (5-way 1-shot).}
  \label{sup:vs:gumtra}
\end{figure}

\begin{figure} [th!]
    \centering
    \includegraphics[width=0.9\linewidth]{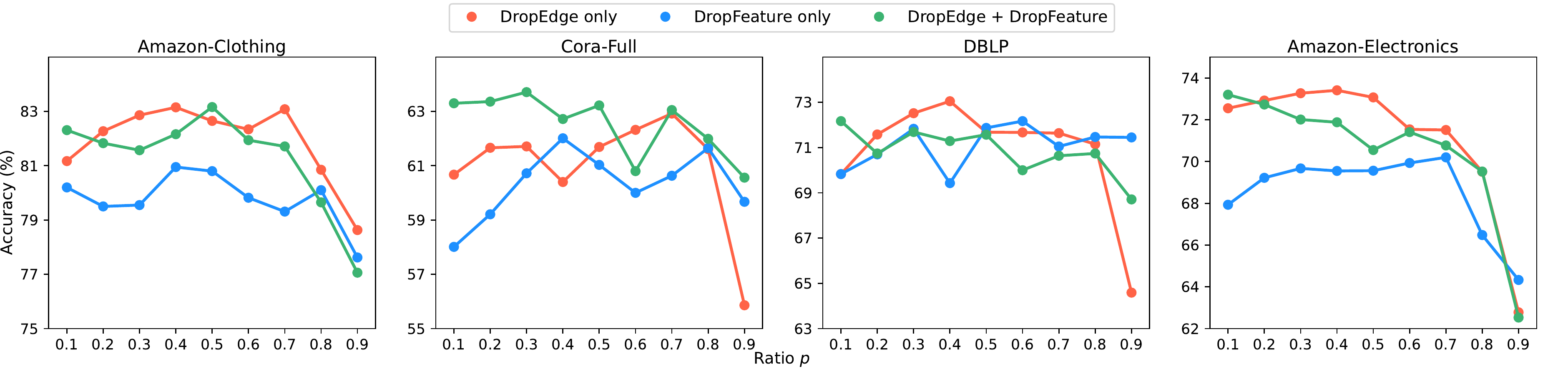}
    \vspace{-3mm}
    \caption{Effect of augmentation function and its strength on g-UMTRA. (5-way 1-shot, base-model: ProtoNet) 
    }
    \label{hyparam:study:gumtra}
\end{figure}

\subsubsection{Drawbacks of g-UMTRA}
Although g-UMTRA can show remarkable performance with some of base-models like ProtoNet~\cite{protonet} (See Figure~\ref{sup:vs:gumtra}), there are several drawbacks of g-UMTRA that limit its applicability in the real-world settings.
First, g-UMTRA requires additional computation of augmented embedding by each update to make query set embeddings, which is time-consuming. Next, g-UMTRA cannot be model-agnostic, since it makes episode within the training phase due to the graph augmentation for the query set generation. Thus, g-UMTRA requires inevitable modification on the training process of some existing models like G-Meta~\cite{gmeta} and TENT~\cite{tent}, up-to-date graph meta-learning methods which are developed under the premise of utilizing supervised episodes, having mutually exclusive support set and query set. Lastly, g-UMTRA is also highly sensitive to the augmentation function choice and its strength (See Figure~\ref{hyparam:study:gumtra}), similar to the original UMTRA.

\subsection{Additional Experimental Results}
\textbf{Impact of the label-scarcity in Cora-Full. }
We additionally conducted the experiment about the label-scarcity problem presented in Figure~\ref{intro:sup_label} in the Cora-Full dataset. Similar to the result shown in Figure~\ref{intro:sup_label}, supervised graph meta-learning methods' FSNC performance decreases as available labeled data and diversity of base classes decrease (See Figure~\ref{appendix:sup_label}).

\textbf{Impact of the label noise in Cora-Full. } 
We also conducted the experiment regarding the label noise presented in Figure~\ref{intro:label_noise} in the Cora-Full dataset. Note that as Cora-Full has smaller size than Amazon-Electronics, we selected label noise $p$ ratio from $\{0, 0.1, 0.2, 0.3\}$. 
As shown in Figure~\ref{appendix:label_noise}, similar to the result in Figure~\ref{intro:label_noise}, supervised meta-learning methods' FSNC performance is highly degraded as noise level increases.

\begin{figure} [bh!]
    \centering
        \subfigure[Impact of the label-scarcity in Cora-Full]{
        \includegraphics[width=0.35\columnwidth]{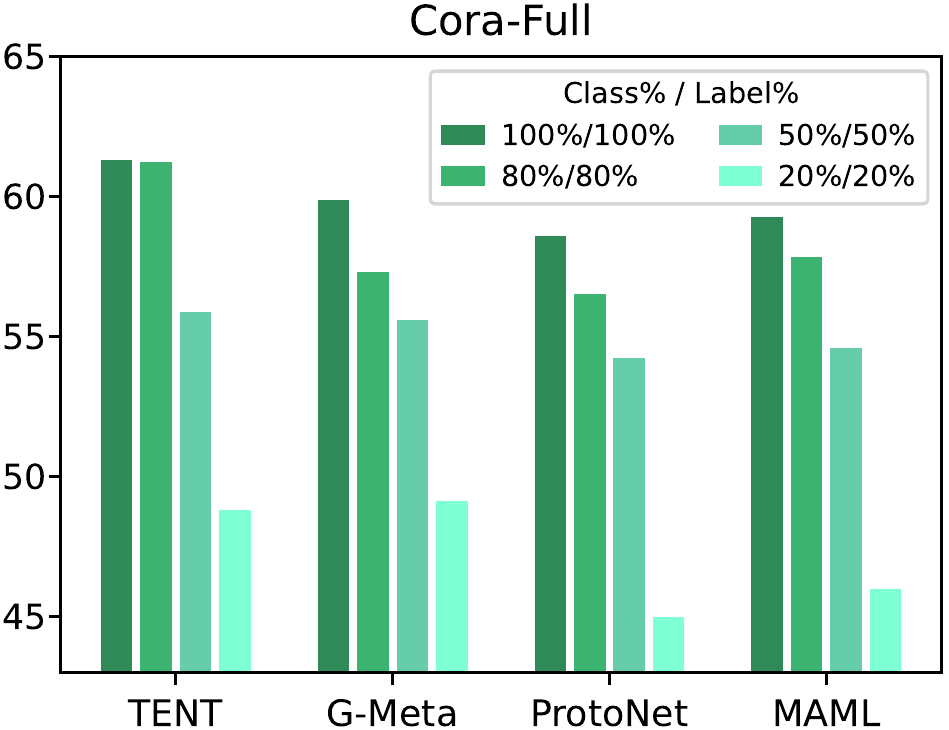}
        \label{appendix:sup_label}
        \vspace{-1mm}
        }
        \hspace{1mm}
        \subfigure[Impact of the label noise in Cora-Full]{
        \includegraphics[width=0.38\columnwidth]{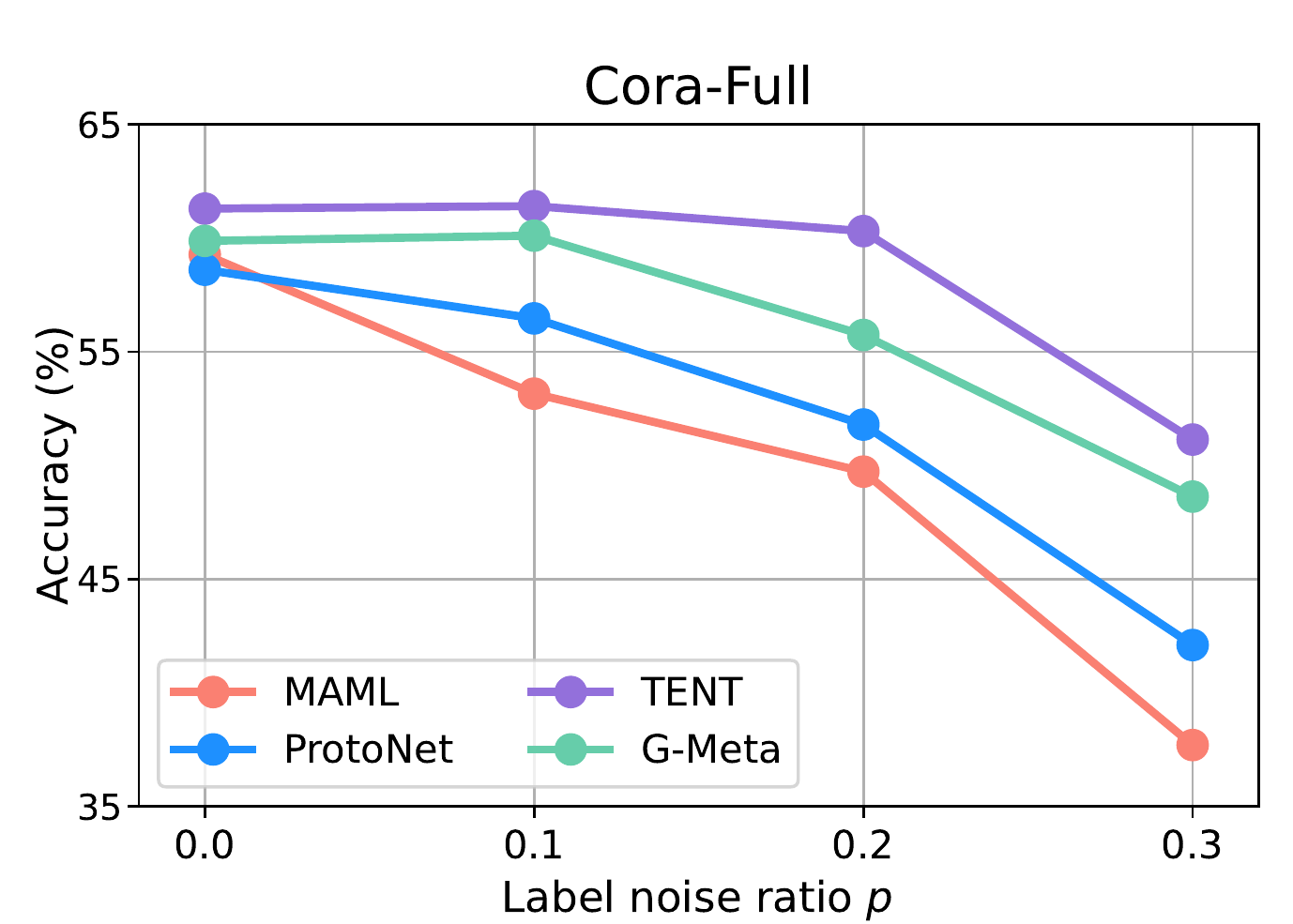}
        \label{appendix:label_noise}
        \vspace{-1mm}
        }
    \vspace{-3mm}
    \caption{(a): Impact of the label-scarcity on supervised graph meta-learning methods, (b): Impact of the (randomly injected) label noise $p$ on supervised graph meta-learning methods. (5-way 1-shot)}
\end{figure}

\clearpage
\begin{table*}[h!]
\centering
\caption{Overall averaged FSNC accuracy (\%) with 95\% confidence intervals on product networks (Full Version)}
\vspace{2mm}
\label{main_table:full:acae}
\renewcommand{\arraystretch}{1.35}
\resizebox{0.9\textwidth}{!}{%
\begin{tabular}{cc|cccc|cccccc}
\hline
\multicolumn{2}{c|}{Dataset} &
  \multicolumn{4}{c|}{\textbf{Amazon Clothing}} &
  \multicolumn{6}{c}{\textbf{Amazon Electronics}} \\ \hline
\multicolumn{2}{c|}{Setting} &
  \multicolumn{2}{c}{5 way} &
  \multicolumn{2}{c|}{10 way} &
  \multicolumn{2}{c}{5 way} &
  \multicolumn{2}{c}{10 way} &
  \multicolumn{2}{c}{20 way} \\ \hline
\multicolumn{1}{c|}{Base Model} &
  Episode Generation &
  1 shot &
  5 shot &
  1 shot &
  5 shot &
  1 shot &
  5 shot &
  1 shot &
  5 shot &
  1 shot &
  5 shot \\ \hline
\multicolumn{1}{c|}{\multirow{3}{*}{MAML}} &
  Supervised &
  76.13±\tiny{1.17} &
  84.28±\tiny{0.87} &
  63.77±\tiny{0.83} &
  76.95±\tiny{0.65} &
  65.58±\tiny{1.26} &
  78.55±\tiny{0.96} &
  57.31±\tiny{0.87} &
  67.56±\tiny{0.73} &
  46.37±\tiny{0.61} &
  60.04±\tiny{0.52} \\
\multicolumn{1}{c|}{} &
  \textbf{\textsc{NaQ-Feat} (Ours)} &
  74.07±\tiny{1.07} &
  86.49±\tiny{0.86} &
  59.44±\tiny{0.91} &
  75.99±\tiny{0.70} &
  59.56±\tiny{1.17} &
  74.85±\tiny{1.03} &
  49.03±\tiny{0.88} &
  70.47±\tiny{0.73} &
  45.27±\tiny{0.60} &
  62.36±\tiny{0.51} \\
\multicolumn{1}{c|}{} &
  \textbf{\textsc{NaQ-Diff} (Ours)} &
  79.30±\tiny{1.17} &
  86.81±\tiny{0.82} &
  69.97±\tiny{0.86} &
  79.74±\tiny{0.68} &
  62.90±\tiny{1.18} &
  78.37±\tiny{0.90} &
  52.23±\tiny{0.84} &
  68.77±\tiny{0.76} &
  43.28±\tiny{0.62} &
  59.88±\tiny{0.51} \\ \hline
\multicolumn{1}{c|}{\multirow{3}{*}{ProtoNet}} &
  Supervised &
  75.52±\tiny{1.12} &
  89.76±\tiny{0.70} &
  65.50±\tiny{0.82} &
  82.23±\tiny{0.62} &
  69.48±\tiny{1.22} &
  84.81±\tiny{0.82} &
  57.67±\tiny{0.85} &
  75.79±\tiny{0.67} &
  48.41±\tiny{0.57} &
  67.31±\tiny{0.47} \\
\multicolumn{1}{c|}{} &
  \textbf{\textsc{NaQ-Feat} (Ours)} &
  83.77±\tiny{0.96} &
  \textbf{92.27±\tiny{0.67}} &
  \underline{76.08±\tiny{0.81}} &
  \underline{85.60±\tiny{0.60}} &
  \textbf{76.46±\tiny{1.11}} &
  \textbf{88.72±\tiny{0.73}} &
  \underline{68.42±\tiny{0.86}} &
  \underline{81.36±\tiny{0.64}} &
  58.80±\tiny{0.60} &
  \textbf{74.60±\tiny{0.47}} \\
\multicolumn{1}{c|}{} &
  \textbf{\textsc{NaQ-Diff} (Ours)} &
  78.64±\tiny{1.05} &
  90.82±\tiny{0.68} &
  71.75±\tiny{0.81} &
  83.81±\tiny{0.60} &
  68.56±\tiny{1.18} &
  84.88±\tiny{0.83} &
  59.46±\tiny{0.86} &
  76.73±\tiny{0.67} &
  49.24±\tiny{0.59} &
  67.99±\tiny{0.48} \\ \hline
\multicolumn{1}{c|}{\multirow{3}{*}{TENT}} &
  Supervised &
  79.46±\tiny{1.10} &
  89.61±\tiny{0.70} &
  69.72±\tiny{0.80} &
  84.74±\tiny{0.59} &
  72.31±\tiny{1.14} &
  85.25±\tiny{0.81} &
  62.13±\tiny{0.83} &
  77.32±\tiny{0.67} &
  52.45±\tiny{0.60} &
  69.39±\tiny{0.50} \\
\multicolumn{1}{c|}{} &
  \textbf{\textsc{NaQ-Feat} (Ours)} &
  \textbf{86.58±\tiny{0.96}} &
  \underline{91.98±\tiny{0.67}} &
  \textbf{79.55±\tiny{0.78}} &
  \textbf{86.10±\tiny{0.60}} &
  \underline{76.26±\tiny{1.11}} &
  \underline{87.27±\tiny{0.81}} &
  \textbf{69.59±\tiny{0.86}} &
  \textbf{81.44±\tiny{0.61}} &
  59.65±\tiny{0.60} &
  \underline{74.09±\tiny{0.46}} \\
\multicolumn{1}{c|}{} &
  \textbf{\textsc{NaQ-Diff} (Ours)} &
  80.87±\tiny{1.08} &
  90.53±\tiny{0.71} &
  72.67±\tiny{0.82} &
  84.54±\tiny{0.61} &
  68.14±\tiny{1.13} &
  83.64±\tiny{0.80} &
  60.44±\tiny{0.79} &
  76.03±\tiny{0.67} &
  51.44±\tiny{0.58} &
  68.37±\tiny{0.49} \\ \hline
\multicolumn{1}{c|}{\multirow{3}{*}{G-Meta}} &
  Supervised &
  78.67±\tiny{1.05} &
  88.79±\tiny{0.76} &
  65.30±\tiny{0.79} &
  80.97±\tiny{0.59} &
  72.26±\tiny{1.16} &
  84.44±\tiny{0.83} &
  61.32±\tiny{0.86} &
  74.92±\tiny{0.71} &
  50.39±\tiny{0.59} &
  65.73±\tiny{0.48} \\
\multicolumn{1}{c|}{} &
  \textbf{\textsc{NaQ-Feat} (Ours)} &
  \underline{85.83±\tiny{1.03}} &
  90.70±\tiny{0.73} &
  73.45±\tiny{0.84} &
  82.61±\tiny{0.66} &
  74.49±\tiny{1.15} &
  84.68±\tiny{0.86} &
  61.18±\tiny{0.83} &
  77.36±\tiny{0.67} &
  55.35±\tiny{0.60} &
  69.16±\tiny{0.51} \\
\multicolumn{1}{c|}{} &
  \textbf{\textsc{NaQ-Diff} (Ours)} &
  82.27±\tiny{1.10} &
  89.88±\tiny{0.77} &
  71.48±\tiny{0.86} &
  82.07±\tiny{0.63} &
  69.62±\tiny{1.20} &
  80.87±\tiny{0.94} &
  58.71±\tiny{0.80} &
  75.55±\tiny{0.67} &
  49.06±\tiny{0.58} &
  67.41±\tiny{0.47} \\ \hline
\multicolumn{1}{c|}{\multirow{3}{*}{GLITTER}} &
  Supervised &
  75.73±\tiny{1.10} &
  89.18±\tiny{0.74} &
  64.30±\tiny{0.79} &
  77.73±\tiny{0.68} &
  66.91±\tiny{1.22} &
  82.59±\tiny{0.83} &
  57.12±\tiny{0.88} &
  76.26±\tiny{0.67} &
  49.23±\tiny{0.57} &
  61.77±\tiny{0.52} \\
\multicolumn{1}{c|}{} &
  \textbf{\textsc{NaQ-Feat} (Ours)} &
  68.24±\tiny{1.27} &
  76.91±\tiny{1.00} &
  59.15±\tiny{0.81} &
  77.19±\tiny{0.65} &
  64.06±\tiny{1.16} &
  80.25±\tiny{0.86} &
  59.31±\tiny{0.79} &
  74.65±\tiny{0.67} &
  49.75±\tiny{0.59} &
  65.30±\tiny{0.51} \\
\multicolumn{1}{c|}{} &
  \textbf{\textsc{NaQ-Diff} (Ours)} &
  70.24±\tiny{1.21} &
  82.48±\tiny{0.83} &
  63.36±\tiny{1.14} &
  80.41±\tiny{0.62} &
  65.45±\tiny{1.22} &
  80.33±\tiny{0.87} &
  54.96±\tiny{0.84} &
  71.10±\tiny{0.72} &
  44.26±\tiny{0.57} & 60.20±\tiny{0.50}
   \\ \hline
\multicolumn{1}{c|}{\multirow{3}{*}{COSMIC}} &
  Supervised &
  82.24±\tiny{0.99} &
  91.22±\tiny{0.73} &
  74.44±\tiny{0.75} &
  81.58±\tiny{0.63} &
  72.61±\tiny{1.05} &
  86.92±\tiny{0.76} &
  65.24±\tiny{0.82} &
  78.00±\tiny{0.64} &
  58.71±\tiny{0.57} &
  70.29±\tiny{0.44} \\
\multicolumn{1}{c|}{} &
  \textbf{\textsc{NaQ-Feat} (Ours)} &
  84.42±\tiny{1.01} &
  91.73±\tiny{0.69} &
  73.15±\tiny{0.78} &
  84.74±\tiny{0.58} &
  73.98±\tiny{1.09} &
  87.08±\tiny{0.75} &
  65.96±\tiny{0.82} &
  79.11±\tiny{0.60} &
  \textbf{61.05±\tiny{0.59}} &
  73.73±\tiny{0.42} \\
\multicolumn{1}{c|}{} &
  \textbf{\textsc{NaQ-Diff} (Ours)} &
  84.40±\tiny{1.01} &
  91.72±\tiny{0.69} &
  73.39±\tiny{0.79} &
  84.82±\tiny{0.58} &
  74.16±\tiny{1.08} &
  87.09±\tiny{0.75} &
  65.95±\tiny{0.81} &
  79.13±\tiny{0.60} &
  \underline{60.40±\tiny{0.59}} &
  73.75±\tiny{0.42} \\ \hline
\multicolumn{1}{c|}{\multirow{3}{*}{TLP}} &
  BGRL &
  81.42±\tiny{1.05} &
  90.53±\tiny{0.71} &
  72.05±\tiny{0.86} &
  83.64±\tiny{0.63} &
  64.20±\tiny{1.10} &
  81.72±\tiny{0.85} &
  53.16±\tiny{0.82} &
  73.70±\tiny{0.66} &
  44.57±\tiny{0.54} &
  65.13±\tiny{0.47} \\
\multicolumn{1}{c|}{} &
  SUGRL &
  63.32±\tiny{1.19} &
  86.35±\tiny{0.78} &
  54.81±\tiny{0.77} &
  73.10±\tiny{0.63} &
  54.76±\tiny{1.06} &
  78.12±\tiny{0.92} &
  46.51±\tiny{0.80} &
  68.41±\tiny{0.71} &
  36.08±\tiny{0.52} &
  57.78±\tiny{0.49} \\
\multicolumn{1}{c|}{} &
  AFGRL &
  78.12±\tiny{1.13} &
  89.82±\tiny{0.73} &
  71.12±\tiny{0.81} &
  83.88±\tiny{0.63} &
  59.07±\tiny{1.07} &
  81.15±\tiny{0.85} &
  50.71±\tiny{0.85} &
  73.87±\tiny{0.66} &
  43.10±\tiny{0.56} &
  65.44±\tiny{0.48} \\ \hline
\multicolumn{2}{c|}{VNT} &
  65.09±\tiny{1.23} &
  85.86±\tiny{0.76} &
  62.43±\tiny{0.81} &
  80.87±\tiny{0.63} &
  56.69±\tiny{1.22} &
  78.02±\tiny{0.97} &
  49.98±\tiny{0.83} &
  70.51±\tiny{0.73} &
  42.10±\tiny{0.53} &
  60.99±\tiny{0.50} \\ \hline
\end{tabular}%
}
\end{table*}

\begin{table*}[h!]
\centering
\caption{Overall averaged FSNC accuracy (\%) with 95\% confidence intervals on product networks (Full Version, OOT: Out Of Time, which means that the training was not finished in 24 hours, OOM: Out Of Memory on NVIDIA RTX A6000)}
\vspace{1mm}
\label{main_table:full:cfdb}
\renewcommand{\arraystretch}{1.4}
\resizebox{\textwidth}{!}{%
\begin{tabular}{cc|cccccc|cccccc}
\hline
\multicolumn{2}{c|}{Dataset} &
  \multicolumn{6}{c|}{\textbf{Cora-full}} &
  \multicolumn{6}{c}{\textbf{DBLP}} \\ \hline
\multicolumn{2}{c|}{Setting} &
  \multicolumn{2}{c}{5 way} &
  \multicolumn{2}{c}{10 way} &
  \multicolumn{2}{c|}{20 way} &
  \multicolumn{2}{c}{5 way} &
  \multicolumn{2}{c}{10 way} &
  \multicolumn{2}{c}{20 way} \\ \hline
\multicolumn{1}{c|}{Base Model} &
  Episode Generation &
  1 shot &
  5 shot &
  1 shot &
  5 shot &
  1 shot &
  5 shot &
  1 shot &
  5 shot &
  1 shot &
  5 shot &
  1 shot &
  5 shot \\ \hline
\multicolumn{1}{c|}{\multirow{3}{*}{MAML}} &
  Supervised &
  59.28±\tiny{1.21} &
  70.30±\tiny{0.99} &
  44.15±\tiny{0.81} &
  57.59±\tiny{0.66} &
  30.99±\tiny{0.43} &
  46.80±\tiny{0.38} &
  72.48±\tiny{1.22} &
  80.30±\tiny{1.03} &
  60.08±\tiny{0.90} &
  69.85±\tiny{0.76} &
  46.12±\tiny{0.53} &
  57.30±\tiny{0.48} \\
\multicolumn{1}{c|}{} &
  \textbf{\textsc{NaQ-Feat} (Ours)} &
  64.64±\tiny{1.16} &
  74.31±\tiny{0.94} &
  49.86±\tiny{0.78} &
  64.88±\tiny{0.64} &
  38.90±\tiny{0.46} &
  53.87±\tiny{0.43} &
  68.49±\tiny{1.23} &
  77.31±\tiny{1.08} &
  55.70±\tiny{0.88} &
  67.94±\tiny{0.82} &
  44.18±\tiny{0.53} &
  56.50±\tiny{0.48} \\
\multicolumn{1}{c|}{} &
  \textbf{\textsc{NaQ-Diff} (Ours)} &
  62.93±\tiny{1.17} &
  76.48±\tiny{0.92} &
  50.10±\tiny{0.83} &
  63.50±\tiny{0.66} &
  38.09±\tiny{0.45} &
  54.08±\tiny{0.41} &
  71.14±\tiny{1.15} &
  79.47±\tiny{1.01} &
  59.18±\tiny{0.91} &
  70.19±\tiny{0.78} &
  44.94±\tiny{0.57} &
  58.68±\tiny{0.47} \\ \hline
\multicolumn{1}{c|}{\multirow{3}{*}{ProtoNet}} &
  Supervised &
  58.61±\tiny{1.21} &
  73.91±\tiny{0.93} &
  44.54±\tiny{0.79} &
  62.15±\tiny{0.64} &
  32.10±\tiny{0.42} &
  50.87±\tiny{0.40} &
  73.80±\tiny{1.20} &
  81.33±\tiny{1.00} &
  61.88±\tiny{0.86} &
  73.02±\tiny{0.74} &
  48.70±\tiny{0.52} &
  62.42±\tiny{0.45} \\
\multicolumn{1}{c|}{} &
  \textbf{\textsc{NaQ-Feat} (Ours)} &
  64.20±\tiny{1.11} &
  79.42±\tiny{0.80} &
  51.78±\tiny{0.75} &
  68.87±\tiny{0.60} &
  40.11±\tiny{0.45} &
  58.48±\tiny{0.40} &
  71.38±\tiny{1.17} &
  82.34±\tiny{0.94} &
  58.41±\tiny{0.86} &
  72.36±\tiny{0.73} &
  47.30±\tiny{0.53} &
  61.61±\tiny{0.46} \\
\multicolumn{1}{c|}{} &
  \textbf{\textsc{NaQ-Diff} (Ours)} &
  65.30±\tiny{1.08} &
  79.66±\tiny{0.79} &
  51.80±\tiny{0.78} &
  \textbf{69.34±\tiny{0.63}} &
  40.76±\tiny{0.49} &
  59.35±\tiny{0.40} &
  73.89±\tiny{1.15} &
  82.24±\tiny{0.98} &
  59.43±\tiny{0.79} &
  72.85±\tiny{0.76} &
  48.17±\tiny{0.52} &
  61.66±\tiny{0.48} \\ \hline
\multicolumn{1}{c|}{\multirow{3}{*}{TENT}} &
  Supervised &
  61.30±\tiny{1.18} &
  77.32±\tiny{0.81} &
  47.30±\tiny{0.80} &
  66.40±\tiny{0.62} &
  36.40±\tiny{0.45} &
  55.77±\tiny{0.39} &
  74.01±\tiny{1.20} &
  \underline{82.54±\tiny{1.00}} &
  \underline{62.95±\tiny{0.85}} &
  \underline{73.26±\tiny{0.77}} &
  49.67±\tiny{0.53} &
  61.87±\tiny{0.47} \\
\multicolumn{1}{c|}{} &
  \textbf{\textsc{NaQ-Feat} (Ours)} &
  64.04±\tiny{1.14} &
  78.48±\tiny{0.79} &
  51.31±\tiny{0.77} &
  67.09±\tiny{0.62} &
  40.04±\tiny{0.48} &
  56.15±\tiny{0.40} &
  72.85±\tiny{1.20} &
  80.91±\tiny{1.00} &
  60.70±\tiny{0.87} &
  71.98±\tiny{0.79} &
  47.29±\tiny{0.53} &
  61.01±\tiny{0.46} \\
\multicolumn{1}{c|}{} &
  \textbf{\textsc{NaQ-Diff} (Ours)} &
  61.85±\tiny{1.12} &
  77.26±\tiny{0.84} &
  49.80±\tiny{0.76} &
  67.65±\tiny{0.63} &
  37.78±\tiny{0.45} &
  56.55±\tiny{0.41} &
  \textbf{76.58±\tiny{1.18}} &
  \textbf{82.86±\tiny{0.95}} &
  \textbf{64.31±\tiny{0.87}} &
  \textbf{74.06±\tiny{0.75}} &
  \textbf{51.62±\tiny{0.54}} &
  63.05±\tiny{0.45} \\ \hline
\multicolumn{1}{c|}{\multirow{3}{*}{G-Meta}} &
  Supervised &
  59.88±\tiny{1.26} &
  75.36±\tiny{0.86} &
  44.34±\tiny{0.80} &
  59.59±\tiny{0.66} &
  33.25±\tiny{0.42} &
  49.00±\tiny{0.39} &
  \underline{74.64±\tiny{1.20}} &
  79.96±\tiny{1.08} &
  61.50±\tiny{0.88} &
  70.33±\tiny{0.77} &
  46.07±\tiny{0.52} &
  58.38±\tiny{0.47} \\
\multicolumn{1}{c|}{} &
  \textbf{\textsc{NaQ-Feat} (Ours)} &
  65.79±\tiny{1.21} &
  79.21±\tiny{0.82} &
  48.90±\tiny{0.80} &
  63.96±\tiny{0.61} &
  40.36±\tiny{0.46} &
  55.17±\tiny{0.43} &
  70.08±\tiny{1.24} &
  80.79±\tiny{0.97} &
  57.98±\tiny{0.87} &
  71.18±\tiny{0.75} &
  45.65±\tiny{0.52} &
  59.38±\tiny{0.46} \\
\multicolumn{1}{c|}{} &
  \textbf{\textsc{NaQ-Diff} (Ours)} &
  62.96±\tiny{1.14} &
  77.31±\tiny{0.87} &
  47.93±\tiny{0.79} &
  63.18±\tiny{0.61} &
  37.55±\tiny{0.46} &
  54.23±\tiny{0.41} &
  70.39±\tiny{1.20} &
  80.47±\tiny{1.03} &
  57.55±\tiny{0.85} &
  69.59±\tiny{0.78} &
  44.56±\tiny{0.52} &
  58.66±\tiny{0.45} \\ \hline
\multicolumn{1}{c|}{\multirow{3}{*}{GLITTER}} &
  Supervised &
  55.17±\tiny{1.18} &
  69.33±\tiny{0.96} &
  42.81±\tiny{0.81} &
  52.76±\tiny{0.68} &
  30.70±\tiny{0.41} &
  40.82±\tiny{0.41} &
  73.50±\tiny{1.25} &
  75.90±\tiny{1.19} &
  OOT &
  OOT &
  OOM &
  OOM \\
\multicolumn{1}{c|}{} &
  \textbf{\textsc{NaQ-Feat} (Ours)} &
  62.66±\tiny{1.12} &
  76.40±\tiny{0.87} &
  50.05±\tiny{0.79} &
  67.66±\tiny{0.61} &
  40.16±\tiny{0.47} &
  57.13±\tiny{0.42} &
  64.55±\tiny{1.18} &
  78.54±\tiny{1.10} &
  OOT &
  OOT &
  OOM &
  OOM \\
\multicolumn{1}{c|}{} &
  \textbf{\textsc{NaQ-Diff} (Ours)} &
  54.58±\tiny{1.14} &
  70.59±\tiny{0.93} &
  47.62±\tiny{0.74} &
  64.58±\tiny{0.65} &
  38.91±\tiny{0.46} &
  52.70±\tiny{0.41} &
  63.44±\tiny{1.21} &
  75.79±\tiny{1.06} &
  OOT &
  OOT &
  OOM &
  OOM \\ \hline
\multicolumn{1}{c|}{\multirow{3}{*}{COSMIC}} &
  Supervised &
  62.24±\tiny{1.15} &
  73.85±\tiny{0.83} &
  47.85±\tiny{0.77} &
  59.11±\tiny{0.60} &
  42.25±\tiny{0.43} &
  47.28±\tiny{0.38} &
  72.34±\tiny{1.18} &
  80.83±\tiny{1.03} &
  59.21±\tiny{0.80} &
  70.67±\tiny{0.71} &
  49.52±\tiny{0.51} &
  59.01±\tiny{0.42} \\
\multicolumn{1}{c|}{} &
  \textbf{\textsc{NaQ-Feat} (Ours)} &
  \textbf{66.30±\tiny{1.15}} &
  \textbf{80.09±\tiny{0.79}} &
  \textbf{52.23±\tiny{0.73}} &
  68.63±\tiny{0.61} &
  \textbf{44.13±\tiny{0.47}} &
  \underline{60.94±\tiny{0.36}} &
  73.55±\tiny{1.16} &
  82.36±\tiny{0.94} &
  58.81±\tiny{0.80} &
  71.14±\tiny{0.70} &
  50.42±\tiny{0.52} &
  \textbf{64.90±\tiny{0.43}} \\
\multicolumn{1}{c|}{} &
  \textbf{\textsc{NaQ-Diff} (Ours)} &
  \underline{66.26±\tiny{1.15}} &
  \underline{80.07±\tiny{0.79}} &
  \underline{52.17±\tiny{0.74}} &
  \underline{68.95±\tiny{0.60}} &
  \underline{44.12±\tiny{0.47}} &
  \textbf{60.97±\tiny{0.37}} &
  73.82±\tiny{1.16} &
  82.29±\tiny{0.94} &
  58.81±\tiny{0.80} &
  71.10±\tiny{0.70} &
  \underline{50.47±\tiny{0.52}} &
  \underline{64.78±\tiny{0.44}} \\ \hline
\multicolumn{1}{c|}{\multirow{3}{*}{TLP}} &
  BGRL &
  62.59±\tiny{1.13} &
  78.80±\tiny{0.80} &
  49.43±\tiny{0.79} &
  67.18±\tiny{0.61} &
  37.63±\tiny{0.44} &
  56.26±\tiny{0.39} &
  73.92±\tiny{1.19} &
  82.42±\tiny{0.95} &
  60.16±\tiny{0.87} &
  72.13±\tiny{0.74} &
  47.00±\tiny{0.53} &
  60.57±\tiny{0.45} \\
\multicolumn{1}{c|}{} &
  SUGRL &
  55.42±\tiny{1.08} &
  76.01±\tiny{0.84} &
  44.66±\tiny{0.74} &
  63.69±\tiny{0.62} &
  34.23±\tiny{0.41} &
  52.76±\tiny{0.40} &
  71.27±\tiny{1.15} &
  81.36±\tiny{1.02} &
  58.85±\tiny{0.81} &
  71.02±\tiny{0.78} &
  45.71±\tiny{0.49} &
  59.77±\tiny{0.45} \\
\multicolumn{1}{c|}{} &
  AFGRL &
  55.24±\tiny{1.02} &
  75.92±\tiny{0.83} &
  44.08±\tiny{0.70} &
  64.42±\tiny{0.62} &
  33.88±\tiny{0.41} &
  53.83±\tiny{0.39} &
  71.18±\tiny{1.17} &
  82.03±\tiny{0.94} &
  58.70±\tiny{0.86} &
  71.14±\tiny{0.75} &
  45.99±\tiny{0.53} &
  60.31±\tiny{0.45} \\ \hline
\multicolumn{2}{c|}{VNT} &
  47.53±\tiny{1.14} &
  69.94±\tiny{0.89} &
  37.79±\tiny{0.69} &
  57.71±\tiny{0.65} &
  28.78±\tiny{0.40} &
  46.86±\tiny{0.40} &
  58.21±\tiny{1.16} &
  76.25±\tiny{1.05} &
  48.75±\tiny{0.81} &
  66.37±\tiny{0.77} &
  40.10±\tiny{0.49} &
  55.15±\tiny{0.46} \\ \hline
\end{tabular}%
}
\end{table*}


\end{document}